\documentclass{article} 
\usepackage{iclr2017_workshop,times}
\usepackage{hyperref}
\usepackage{url}
\usepackage{tabularx}
\usepackage{latexsym}
\usepackage{graphicx}
\usepackage{float}
\usepackage{color}
\usepackage{amsmath}
\usepackage{amssymb}
\usepackage{subfig}
\usepackage{multirow}
\usepackage{array}
\usepackage{tikz}
\usepackage{pgfplotstable}
\usetikzlibrary{arrows, patterns,backgrounds,fit,positioning}

\title{Gated Multimodal Units for Information Fusion}
\author{Arevalo, John \\
Dept. of Computing Systems and Industrial Engineering\\
Universidad Nacional de Colombia\\
Cra 30 No 45 03-Ciudad Universitaria\\
\texttt{jearevaloo@unal.edu.co} \\
\And
Solorio, Thamar\\
Dept. of Computer Science\\
University of Houston\\
Houston, TX 77204-3010\\
\texttt{solorio@cs.uh.edu}\\
\And
Montes-y-G\'{o}mez, Manuel \\
Instituto Nacional de Astrof\'{i}sica, \'{O}ptica y Electr\'{o}nica\\
Computer Science Department\\
Luis Enrique Erro No. 1, Sta. Ma. Tonantzintla\\
C.P. 72840 Puebla, Mexico\\
\texttt{smmontesg@inaoep.mx}\\
\And
Gonz\'{a}lez, Fabio A. \\
Dept. of Computing Systems and Industrial Engineering\\
Universidad Nacional de Colombia\\
Cra 30 No 45 03-Ciudad Universitaria\\
\texttt{fagonzalezo@unal.edu.co} \\
}

\begin{document}

\maketitle

\begin{abstract}
This paper presents a novel model for multimodal learning based on gated neural networks. The Gated Multimodal Unit (GMU) model is intended to be used as an internal unit in a neural network architecture whose purpose is to find an intermediate representation based on a combination of data from different modalities. The GMU learns to decide how modalities influence the activation of the unit using multiplicative gates. It was evaluated on a multilabel scenario for genre classification of movies using the plot and the poster. The GMU improved the macro f-score performance of single-modality approaches and outperformed other fusion strategies, including mixture of experts models. Along with this work, the MM-IMDb dataset is released which, to the best of our knowledge, is the largest publicly available multimodal dataset for genre prediction on movies.
\end{abstract}
\section{Introduction}

Representation learning methods have received a lot of attention by researchers and practitioners because of its successful application to complex problems in areas such as computer vision, speech recognition and text processing \citep{lecun2015deep}. Most of these efforts have concentrated on data involving one type of information (images, text, speech, etc.), despite data being naturally multimodal. Multimodality refers to the fact that the same real-world concept can be described by different views or data types. Collaborative encyclopedias (such as Wikipedia) describe a famous person through a mixture of text, images and, in some cases, audio. Users from social networks comment events like concerts or sport games with small phrases and multimedia attachments (images/videos/audios). Medical records are represented by a collection of images, sound, text and signals, among others. The increasing availability of multimodal databases from different sources has motivated the development of automatic analysis techniques to exploit the potential of these data as a source of knowledge in the form of patterns and structures that reveal complex relationships \citep{bhatt2011multimedia, atrey2010multimodal}. In recent years, multimodal tasks have acquired attention by the representation learning community. Strategies for visual question answering \citep{antol2015vqa}, or image captioning \citep{vinyals2015show,xu2015show,johnson2015densecap} have developed interesting ways of combining different representation learning architectures.

Most of these models are focused on mapping from one modality to another or solving an auxiliary task to create a common representation with the information of all modalities. In this work, we design a novel module that combines multiple sources of information, which is optimized with respect to the end goal objective function. Our proposed module is based on the idea of gates for selecting which parts of the input are more likely to contribute for correctly generating the desired output. We use multiplicative gates that assign importance to various features simultaneously, creating a rich multimodal representation that does not require manual tuning, but instead it learns directly from the training data. Our gated model can be reused in different network architectures for solving different tasks, and can be optimized end-to-end with other modules in the architecture using standard gradient-based optimization algorithms.

\begin{figure}
\centering{}\includegraphics[width=1\columnwidth]{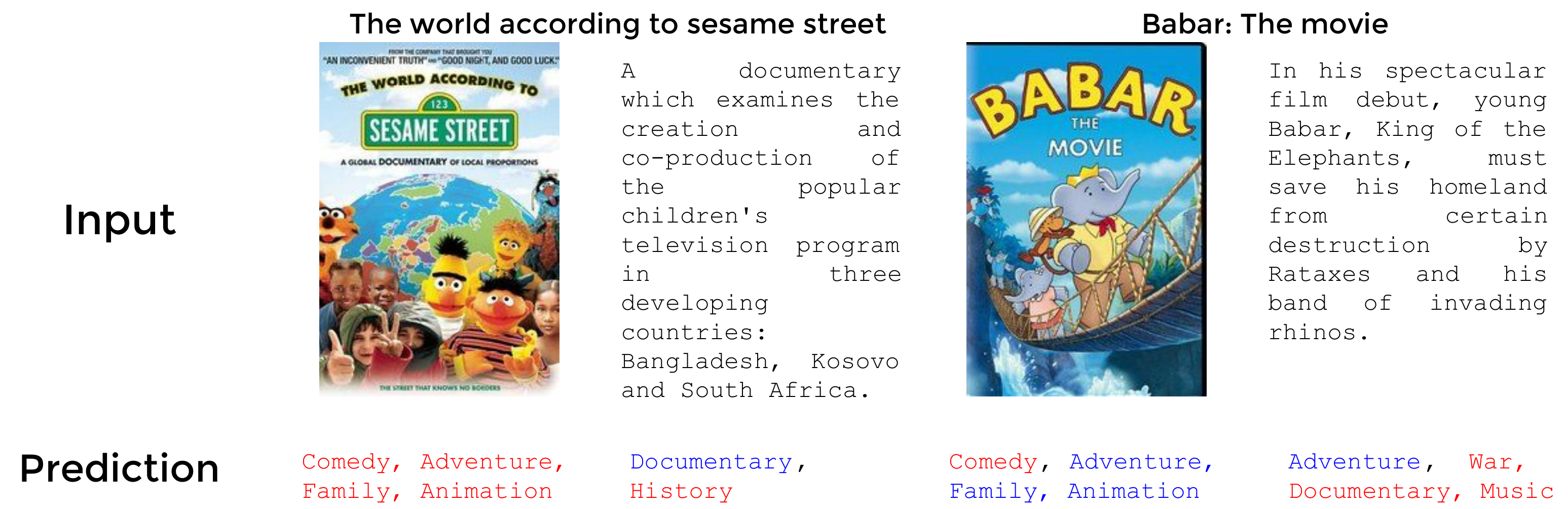}
\caption{Predictions of genre labels depending on the input modality. Red and blue labels indicate false positives and true positives respectively.}
\label{fig:overall}
\end{figure}

As an application use case, we explore the task of identifying a movie genre based on its plot and its poster. Genre classification has several application areas like document categorization \citep{kanaris2009learning}, recommendation systems \citep{makita2016movie}, and information retrieval systems, among others. Figure \ref{fig:overall} depicts the challenging task of assigning genres to a particular movie based solely on the usage of one modality. Such predictions were done with \emph{MaxoutMLP\_w2v} and \emph{VGG\_transfer} approaches (See Section \ref{sec:method}), both of them are models based on representation learning. It can be seen that even a human might be confused if both modalities are not available. The main hypothesis of this work is that a model using gating units, in contrast to a hand-coded multimodal fusion architecture, will be able to learn an input-dependent gate-activation pattern that determines how each modality contribute to the output of hidden units.

The rest of the paper is organized as follows: Section \ref{sec:rel_work} presents a literature review and some considerations of the previous work. Section \ref{sec:method} describes the methods used as baseline as well as our representation-leaning-based model proposed. Section \ref{sec:evaluation} presents the experimental evaluation setup along with the details of the MM-IMDb dataset. Section \ref{sec:results} shows and discusses the results for movie genre classification. Finally, Section \ref{sec:conclusions} draws the conclusions and future work.

\section{Related work}\label{sec:rel_work}

\subsection{Multimodal fusion}
Different reviews \citep{atrey2010multimodal,bhatt2011multimedia, deng2014deep,deng2014tutorial} have summarized strategies that addressed multimodal analysis. Most of the collected works claimed the superiority of multimodal over unimodal approaches for automatic analysis tasks.  A conventional multimodal analysis system receives as input two or more modalities that describe a particular concept. The most common multimodal sources are video, audio, images and text. In recent years there has been a consensus with respect to the use of representation learning models to characterize the information of this kind of sources \citep{lecun2015deep}. However, the way that such extracted features are combined is still in exploration.

Multimodal combination seeks to generate a single representation that makes easier automatic analysis tasks when building classifiers or other predictors. A simple approach is to concatenate features to get a final representation \citep{kiela2014learning, pei2013unsupervised, suk2013deep}. Although it is a straightforward strategy, it ignores inherent correlations between different modalities.

More complex fusion strategies include Restricted Boltzmann Machines (RBMs) and autoencoders. \citet{ngiam2011multimodal} concatenated higher level representations and train two RBMs to reconstruct the original audio and video representations respectively. Additionally, they trained a model to reconstruct both modalities given only one of them as input. In an interesting result, \citet{ngiam2011multimodal} were able to mimic a perceptual phenomenon that demonstrates an interaction between hearing and vision in speech perception known as McGurk effect. A similar approach was proposed by \citet{srivastava2012multimodal}. They modified feature learning and reconstruction phases with Deep Boltzmann Machines. Authors claimed that such strategy is able to exploit large amounts of unlabeled data by improving the performance in retrieval and annotation tasks. Other similar strategies propose to fusion modalities using neural network architectures \citep{andrew2013deep,feng2013constructing,kang2012deep,kiros2013multimodal,lu2014learning,mao2014explain,tu2014multimedia,wu2013online} with two input layers separately and including a final supervised layer such as softmax regression classifier.

An alternative approach involves an objective or loss function suited for the target task \citep{akata2014zero, frome2013devise, kiros2014unifying, mao2014explain, socher2013zero, socher2014grounded, zheng2014topic}. These strategies usually assume that there exists a common latent space where modalities can express the same semantic concept through a set of transformations of the raw data. The semantic embedding representations are such that two concepts are similar if and only if their semantic embeddings are close \citep{norouzi2014zero}. In \citep{socher2013zero} a multimodal strategy to perform zero-shot classification was proposed. They trained a word-based neural network model \citep{huang2012improving} to represent textual information, whilst use unsupervised feature learning models proposed in \citep{coates2011importance} to get image representation. The fusion was done by learning an image linear mapping to project images into the semantic word space learned in the neural network model. Additionally a Bayesian framework was included to decide whether an image is of a seen or unseen class. \citet{frome2013devise} learn the image representation using a CNN trained with the Imagenet dataset and a word-based neural language model \citep{mikolov2013distributed} to represent the textual modality. To perform the fusion they re-train CNN using text representation as targets. This work outperforms scalability with respect to \citep{socher2013zero} from 2 to 20,000 unknown classes in the zero-shot learning task. A modified strategy of \citet{frome2013devise} was presented by \citet{norouzi2014zero}. Instead of re-train the CNN network, they built a convex combination with probabilities estimated by the classifier and semantic embedding vector of the unseen label. This simple strategy outperforms state-of-the-art results. Because the cost function involves both multimodal combination and supervision, these family of models are tied to the task of interest. Thus, if the domain or task conditions changes, adaptations are required.

The proposed model is closely related to the mixture of experts (MoE) approach \citep{jacobs1991adaptive}. However, the common usage of MoE is focused on performing decision fusion, i.e. combining predictors to address a supervised learning problem \citep{yuksel2012twenty}. Our model is devised as a new component in the representation learning scheme, making it independent from the final task (e.g. classification, regression, unsupervised learning, etc) provided that the defined cost function be differentiable. 

\subsection{Movie genre classification}
With respect to movie genre classification, several strategies also have been proposed. These strategies have used different modalities to characterize each movie, such as textual features, image features and multimedia features (audio and/or video). \citet{huang2007film} were one of the first teams exploring this task. They classified movie previews into 3 genres by extracting handcrafted features from the video and training a decision tree classifier. They evaluated the model using 44 films. Using only textual modality, \citet{shah2013movie} performed single-label genre classification of movie scripts using clustering algorithms with 260 movies. Later, combining two modalities, \citet{pais2012animated} classified movies between drama and non-drama using visual and textual features with 107 samples. \citet{hong2015multimodal} explored different PLSA models to combine 3 modalities: audio, image and text to predict genre of movie previews. It was single label classification with 4 genres for 140 movies taken from IMDb.

Recently, \citet{fu2015fast} used a set of handcrafted visual features for poster characterization and bag-of-words for synopsis. Then, they trained one SVM per each modality to combine their predictions. The dataset contained $2,400$ movies with one genre (out of 4) each.

The previous mentioned works present this problem in a single label setup. However, a more realistic scenario would be multilabel, since most of the movies belong to more than one genre, (e.g. Matrix(2000) is a Sci-fi/Action movie). In this setup, \citet{anand2014evaluating} explores the efficiency of using keywords and users' tags to perform multilabeling using the movies from MovieLens 1M dataset which contains $1,700$ movies. Also \citet{ivasic-kos2014movie,ivasic-kos2015automatic} performed multilabel classification using handcrafted features from posters, with $1,500$ samples for 6 genres. \citet{makita2016movie,makita2016multinomial} use movie ratings matrix and genre correlation matrix to predict the genre. It used a smaller version of the Movielens dataset with 18 movie genres.

Most of the above works have used the publicly available MovieLens datasets. However, there is not a single experimental setup defined so that all methods can be systematically compared. Also, to the best of our knowledge, none of the previous works contain more than $10,000$ samples. With this work we will release a dataset created with the movies of the MovieLens 20M dataset. We include not only genre, poster and plot information used in this work, but also the poster of the movie as well as more than 50 characteristics taken from the IMDb website. We will also release the source code to automatically add more movies and genres.

\section{Methods}\label{sec:method}
This paper presents a neural-network-based strategy for multilabel classification of multimodal data. The key component of the strategy is a novel type of hidden unit, the Gated Multimodal Unit (GMU), which learns to decide how modalities influence the activation of the unit using gates. The details of the GMU are presented in Subsection \ref{subsec:GMU}. 

Statistical properties usually are not shared across modalities \citep{srivastava2012multimodal}. And thus, they require different representation strategies according to the nature of data. This work explored several strategies to address text and visual representation. For text information we evaluated word2vec models, n-grams models and RNN models. The details are discussed in Subsection \ref{subsec:text}. On the other hand, two different convolutional neural networks were evaluated for processing visual data and are presented in Subsection \ref{subsec:visual}.

\subsection{Gated multimodal unit for multimodal fusion}\label{subsec:GMU}
Multimodal learning is closely related to data fusion. Data fusion looks for optimal ways of combining different information sources into an integrated representation that provides more information than the individual sources \citep{bhatt2011multimedia}. This fusion can be performed at different levels, that can be categorized into two broad categories: feature fusion and decision fusion. Feature fusion, also called early fusion, looks for a subset of features from different modalities, or combinations of them, that better represent the information needed to solve a particular problem. On the other hand, decision fusion, or late fusion, combines decisions from different systems, e.g. classifiers, to produce consensus. This consensus may be reached by a simple average, a voting system or a more complex Bayesian framework.

\begin{figure}
    \subfloat[]{\begin{tikzpicture}[
cross/.style={path picture={
  \draw[black]
(path picture bounding box.south east) -- (path picture bounding box.north west) (path picture bounding box.south west) -- (path picture bounding box.north east);
}},
conn/.style={->, rounded corners, line width=0.60mm, -latex},
elemwise/.style={circle, fill=wise, draw, thick},
act/.style={rectangle, fill=act, draw, thick},
bg/.style={fill=bgcolor, draw=bgline, line width=0.50mm, rounded corners}
]
	\definecolor{bgcolor}{RGB}{225, 247, 208};
	\definecolor{bgline}{RGB}{109, 135, 89};
    \definecolor{act}{RGB}{245, 238, 156};
    \definecolor{wise}{RGB}{249, 209, 209};
    \node (x1) at (0, -1) {$x_1$};
    \node[act] (z1) at (1, 1) {$\sigma$};
    \node[act] (h1) at (0, 2) {$\tanh$};
    \node[elemwise, cross] (g1) at (0, 3) {};
    \node (x2) at (2, -1) {$x_2$};
    \node[act] (z2) at (3, 1) {$\sigma$};
    \node[act] (h2) at (2, 2) {$\tanh$};
    \node[elemwise, cross] (g2) at (2, 3) {};
    \node[scale=2] (dots) at (4, 2) {${\cdots}$};
    \node (xk) at (5, -1) {$x_k$};
    \node[act] (zk) at (6, 1) {$\sigma$};
    \node[act] (hk) at (5, 2) {$\tanh$};
    \node[elemwise, cross] (gk) at (5, 3) {};
    \node[elemwise] (sum) at (3, 4) {$+$};
    \node (pcon) at (2, 0) {};
    \begin{pgfonlayer}{background}
    \node[fit=(z1) (sum) (h1) (zk) (pcon), bg] (container) {};
    \end{pgfonlayer}
    \node (h) at (3, 5) {$h$};
    \draw[conn] (x1) -- (h1);
    \draw[conn] (z1) |- (g1) node[left, pos=0.125] {$z_1$};
    \draw[conn] (h1) -- (g1) node[left, pos=0.5] {$h_1$};
    \draw[conn] (g1) |- (sum);
    
    \draw[conn] (x2) -- (h2);
    \draw[conn] (z2) |- (g2) node[left, pos=0.125] {$z_2$};
    \draw[conn] (h2) -- (g2) node[left, pos=0.5] {$h_2$};
    \draw[conn] (g2) |- (sum);
    \draw[conn] (xk) -- (hk);
    \draw[conn] (zk) |- (gk) node[left, pos=0.125] {$z_k$};
    \draw[conn] (hk) -- (gk) node[left, pos=0.5] {$h_k$};
    \draw[conn] (gk) |- (sum);
    \draw[conn] (sum) -- (h);
    \draw[bgcolor,line width=2.5mm] (1,0) -- (6, 0);
    \draw[conn] (x1) -| ++(1, 1) -| (z1);
    \draw[conn] (x1) -| ++(1, 1) -| (z2);
    \draw[conn] (x1) -| ++(1, 1) -| (zk);
    \draw[conn] (x2) -| ++(1, 1) -| (z1);
    \draw[conn] (x2) -| ++(1, 1) -| (z2);
    \draw[conn] (x2) -| ++(1, 1) -| (zk);
    \draw[conn] (xk) -| ++(1, 1) -| (z1);
    \draw[conn] (xk) -| ++(1, 1) -| (z2);
    \draw[conn] (xk) -| ++(1, 1) -| (zk);
  \end{tikzpicture}}\hfill
    \subfloat[]{\begin{tikzpicture}[
cross/.style={path picture={
  \draw[black]
(path picture bounding box.south east) -- (path picture bounding box.north west) (path picture bounding box.south west) -- (path picture bounding box.north east);
}},
conn/.style={->, rounded corners, line width=0.60mm, -latex},
elemwise/.style={circle, fill=wise, draw, thick},
act/.style={rectangle, fill=act, draw, thick},
bg/.style={fill=bg, draw=bgline, line width=0.50mm, rounded corners}
]
	\definecolor{bg}{RGB}{225, 247, 208};
	\definecolor{bgline}{RGB}{109, 135, 89};
    \definecolor{act}{RGB}{245, 238, 156};
    \definecolor{wise}{RGB}{249, 209, 209};
    \node (xv) at (0.5, 0) {$x_v$};
    \node (xt) at (4, 0) {$x_t$};
    \node[act] (g) at (2, 1) {$\sigma$};
    \node[act] (hv) at (0.5, 2) {$\tanh$};
    \node[act] (ht) at (4, 2) {$\tanh$};
    \node[elemwise, cross] (gv) at (0.5, 3) {};
    \node[elemwise, cross] (gt) at (4, 3) {};
    \node[elemwise] (neg) at (3, 3) {\small $1-$};
    \node[elemwise] (sum) at (2, 4) {$+$};
    \begin{pgfonlayer}{background}
    \node[fit=(g) (sum) (ht) (hv), bg] (container) {};
    \end{pgfonlayer}
    \node (h) at (2, 5) {$h$};
    \draw[conn] (xv) |- (g);
    \draw[conn] (xt) |- (g);
    \draw[conn] (xv) -- (hv);
    \draw[conn] (xt) -- (ht);
    \draw[conn] (g) |- (neg); 
    \draw[conn] (g) |- (gv) node[left, pos=0.125] {$z$};
    \draw[conn] (neg) -- (gt);
    \draw[conn] (ht) -- (gt) node[left, pos=0.5] {$h_t$};
    \draw[conn] (hv) -- (gv) node[left, pos=0.5] {$h_v$};
    \draw[conn] (gv) |- (sum);
    \draw[conn] (gt) |- (sum);
    \draw[conn] (sum) -- (h);
  \end{tikzpicture}}
    \caption{Illustration of gated units. a) The proposed model to use with more than two modalities. b) A simplification for the bimodal approach.}
    \label{fig:gmu}
\end{figure}
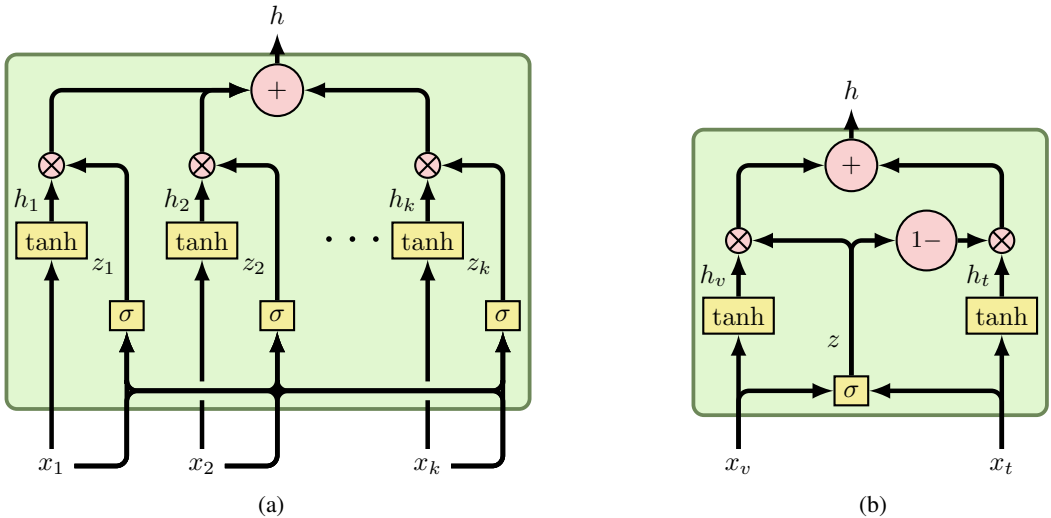

In this work we present a model, based on gated neural networks, for data fusion that combines ideas from both feature and decision fusion. The model, called Gated Multimodal Unit (GMU), is inspired by the flow control in recurrent architectures like GRU or LSTM. A GMU is intended to be used as an internal unit in a neural network architecture whose purpose is to find an intermediate representation based on a combination of data from different modalities. Figure \ref{fig:gmu}.a depicts the structure of a GMU. Each $x_i$ corresponds to a feature vector associated with modality $i$. Each feature vector feeds a neuron with a $\tanh$ activation function, which is intended to encode an internal representation feature based on the particular modality. For each input modality, $x_i$, there is a gate neuron (represented by $\sigma$ nodes in the diagram), which controls the contribution of the feature calculated from $x_i$ to the overall output of the unit. When a new sample is fed to the network, a gate neuron associated to modality $i$ receives as input the feature vectors from all the modalities and uses them to decide whether the modality $i$ may contribute, or not, to the internal encoding of the particular input sample.

Figure \ref{fig:gmu}.b shows a simplified version of the GMU for two input modalities, $x_v$ (visual modality) and $x_t$ (textual modality), that will be used in the remaining of the paper. It should be noted that both models are not completely equivalent, since in the bimodal case the gates are tied. Such weight tying constraints the model, so that the units trade off between both modalities while they use less parameters than the multimodal case. The equations governing this GMU are as follows:
\begin{align*}
h_v &= \tanh\left(W_v\cdot x_v\right)\\
h_t &= \tanh\left(W_t\cdot x_t\right)\\
z &= \sigma\left(W_z \cdot \left[x_v, x_t\right] \right)\\
h &= z * h_v + \left(1 - z \right) * h_t\\
\Theta &= \left\{W_v, W_t,W_z\right\}
\end{align*}

with $\Theta$ the parameters to be learned and $\left[\cdot,\cdot\right]$ the concatenation operator. Since all are differentiable operations, this model can be easily coupled with other neural network architectures and trained with stochastic gradient descent.

\subsection{Text representation}\label{subsec:text}

Text representation is a critical step when classification tasks are addressed using machine learning methods. Traditional approaches are based on counting frequencies of $n$-gram occurrences such as words or sequences of characters (e.g. bag-of-words models). The main drawback of such approaches is the difficulty to model relationships between words and their context. An alternative approach was initially proposed by \citet{bengio2003neural}, by building a language model based on a neural network architecture (NNLM). The NNLM was able to learn distributed representations of words that capture contextual information. Later, this model was simplified to deal with large corpora by removing hidden layers in the neural network architecture (word2vec) \citep{mikolov2013efficient}. This is a fully unsupervised model that takes advantage of large sets of unlabeled documents. Herein, three text representations were evaluated:

\begin{description}
\item[{n-gram}] Following the strategy proposed by \citet{kanaris2009learning}, we used the n-gram strategy for representing text. Despite their simplicity, $n$-gram models have shown to be a competitive baseline. 

\item [{Word2Vec}] Word2vec is an unsupervised learning algorithm that finds a vector representation for each word based on its context \citep{mikolov2013efficient}. It has been shown that this model is able to find semantic and syntactic relationships using arithmetic operations between the vectors. Based on this property, we represent a movie as the average of the vectors of words in the plot outline. The main motivation to aggregate word2vec vectors is the property of additive compositionality that this representation has exposed over different set of tasks such as word analogies. The usual way to aggregate is to sum vectors. We instead take the average to avoid large input values to the neural network.

\item [{Recurrent neural network}] Here we take the plot outline as a sequence of words and train a supervised recurrent neural network. We evaluated two variants. The first one (\emph{RNN\_w2v}) is a transfer learning model that takes as input the word vectors of word2vec as representations. The second one learns the word vectors from scratch (\emph{RNN\_end2end}).

\end{description}
\subsection{Visual representation}\label{subsec:visual}
In computer vision tasks, Convolutional neural networks have become the \emph{de facto} standard. It has been shown that CNN models trained with a huge amount of data are able to learn common features shared across different domains. This characteristic is usually exploited by transfer learning approaches. For visual representation we explored 2 strategies: transfer learning and end-to-end training.
\begin{description}
\item [{VGG Transfer}] In this approach, the VGG Network \citep{simonyan2014very} trained with the ImageNet dataset is used as feature extractor by taking the last hidden activations as the visual representation.
\item [{End2End CNN}] Here, a CNN with 5 convolutional layers and an MLP (see Section \ref{sub:classification}) on top was trained from scratch.
\end{description}

\subsection{Classification model}\label{sub:classification}

Based on the defined representation, we explored two methods to map from feature vectors to genre classification. In particular we explored a simple Logistic regression and a neural network architecture. This is a multilayer perceptron (MLP) with two fully connected layers and maxout activation function. In particular, the maxout activation function $h_i:\mathbb{R}^{n}\rightarrow\mathbb{R}$ is a defined as: 
\begin{equation}
h_{i}\left(\mathbf{s}\right)=\underset{j\in\left[1,k\right]}{\mbox{max}}z_{i,j}
\end{equation}

where $\mathbf{s}\in\mathbb{R}^{n}$ is the input vector, $z_{i,j}=\mathbf{s}^{T}\mathbf{W}_{\cdots ij}+\mathbf{b}_{ij}$ is the output of the $j$-th linear transformation of the $i$-th hidden unit, and $\mathbf{W}\in\mathbb{R}^{d\times m\times k}$ and $\mathbf{b}\in\mathbb{R}^{m\times k}$ are learned parameters. It has been shown that maxout models with just $2$ hidden units behave as universal approximators, while are less prone to saturate units \citep{goodfellow2013maxout}.

\section{Experimental evaluation}
\label{sec:evaluation}

\subsection{Multimodal IMDb dataset}\label{sec:dataset}
\begin{figure}
\centering
\includegraphics[width=0.6\textwidth]{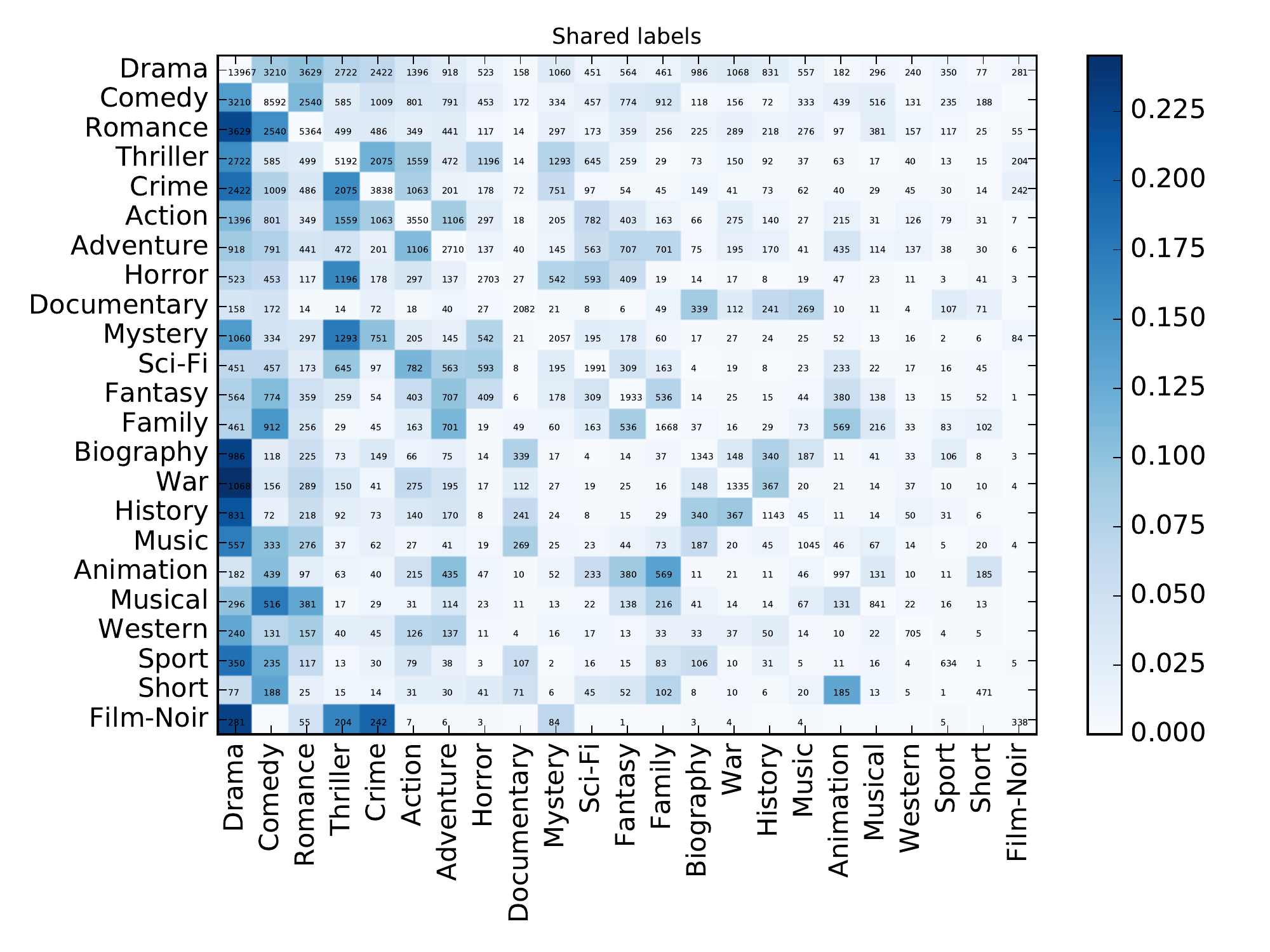}\caption{Co-ocurrence matrix of genre tags\label{fig:Coocurrence-matrix}}
\end{figure}
With this work we will make publicly available the Multimodal IMDb (\textbf{MM-IMDb})\footnote{\url{http://lisi1.unal.edu.co/mmimdb/}} dataset. MM-IMDb dataset is built with the IMDb id's provided by the Movielens 20M dataset \footnote{\url{http://grouplens.org/datasets/movielens/}} that contains ratings of $27,000$ movies. Using the IMDbPY  \footnote{\url{http://imdbpy.sourceforge.net/}} library, movies which do not contain their poster image were filtered out. As the final result, the MM-IMDb dataset comprises $25,959$ movies along with their plot, poster, genres and other 50 additional metadata fields such as year, language, writer, director, aspect ratio, etc.

\begin{figure}
\begin{minipage}{.5\textwidth}
\centering{}\includegraphics[width=\textwidth]{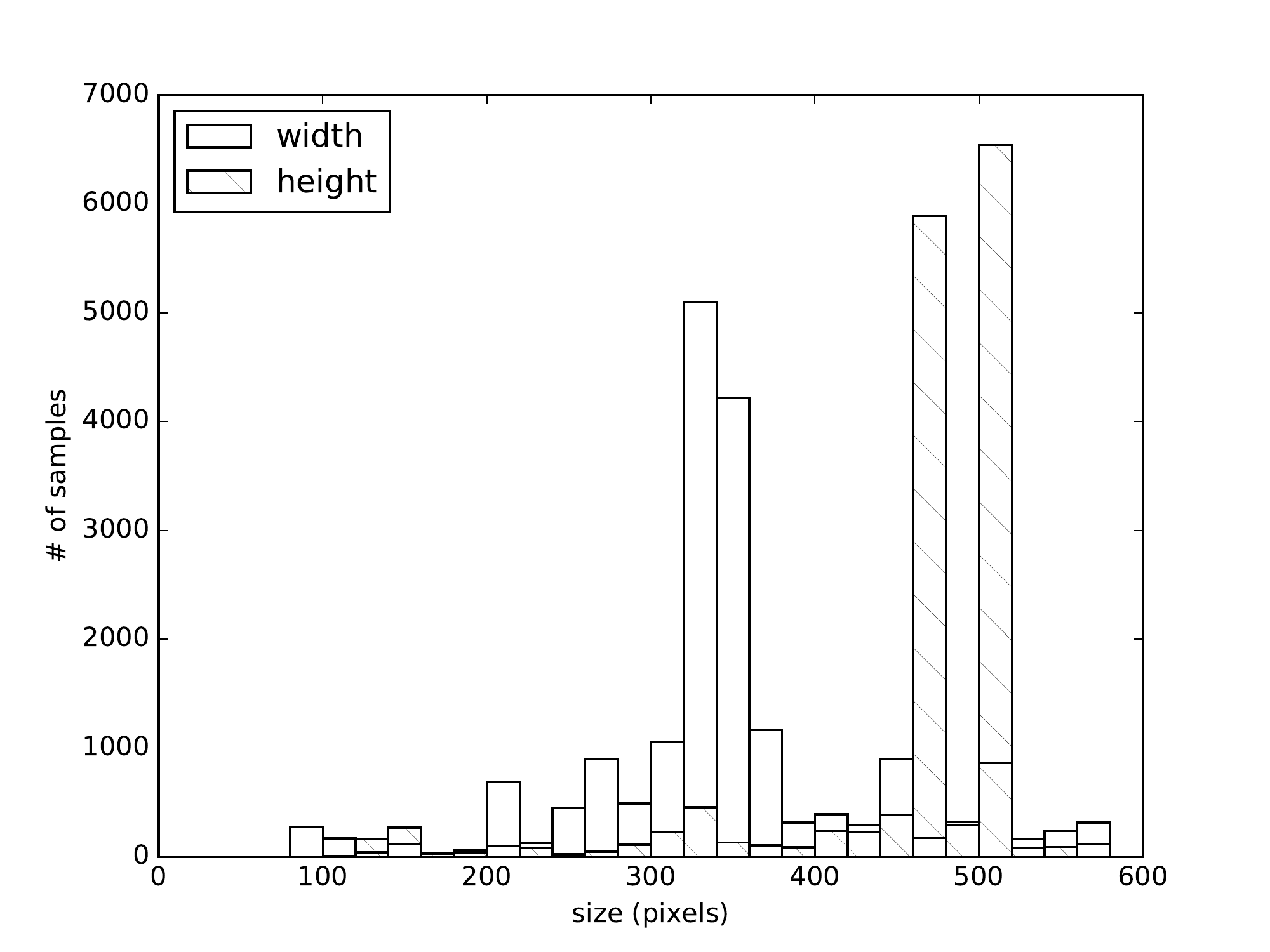}\caption{Size distribution of movie posters. \label{fig:pixel_dist}}
\end{minipage}
\begin{minipage}{.5\textwidth}
\centering{}\includegraphics[width=\textwidth]{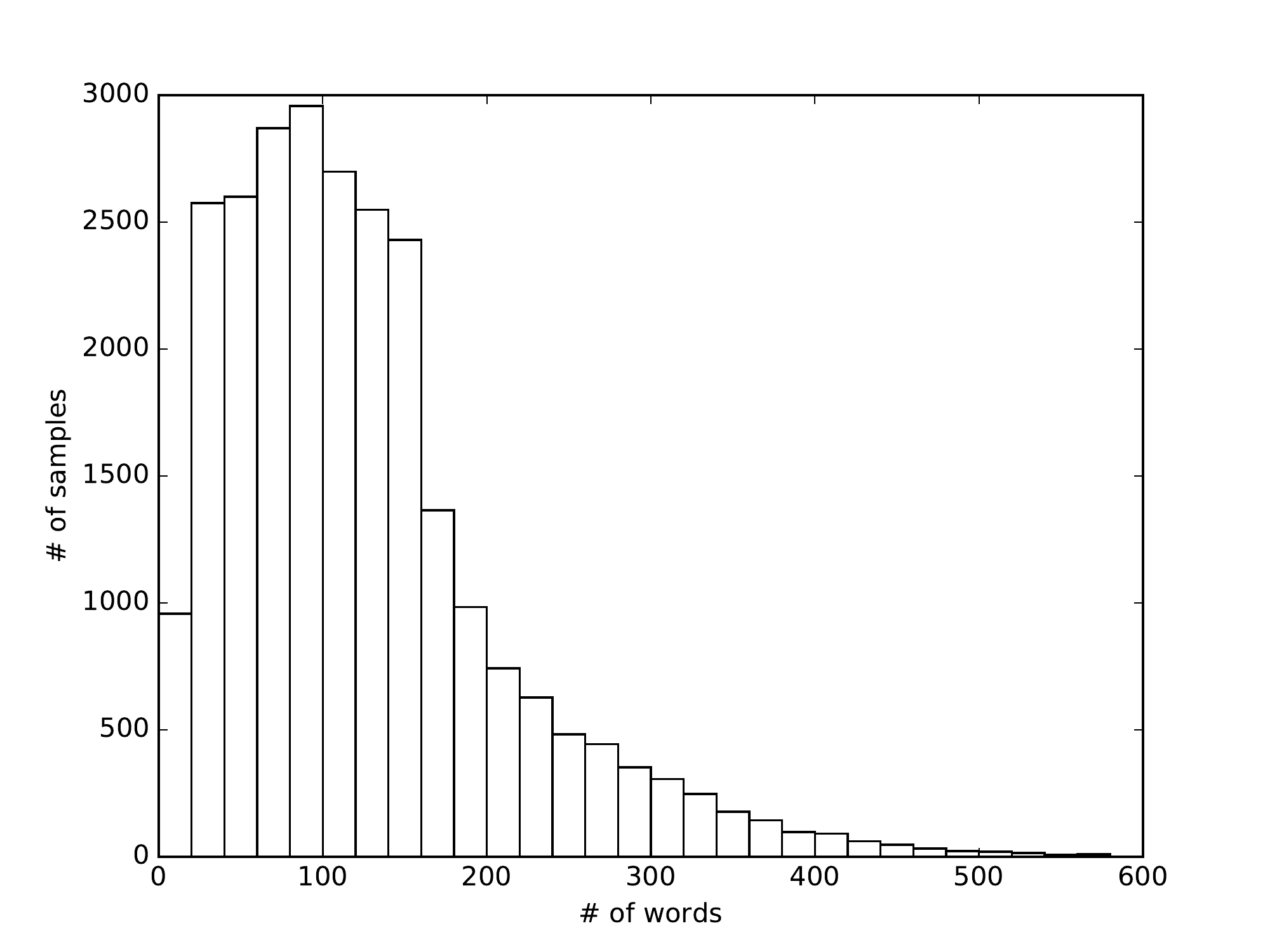}\caption{Length distribution of movie plots. \label{fig:length_dist}}
\end{minipage}
\end{figure}

Notice that one movie may belong to more than one genre. Figure \ref{fig:Coocurrence-matrix} shows the co-occurrence matrix, where the color bar indicates the representative co-occurrence per row, while Figure \ref{fig:pixel_dist} and Figure \ref{fig:length_dist} depict the distribution of the movie poster sizes and length of movie plots respectively. Each plot contains on average 92.5 words, while the longest one contains $1,431$ words and the average of genres per movie is $2.48$. In this work, we defined the task of movie genre prediction based on its plot and image poster. Nevertheless, the additional metadata information encourages other interesting tasks such as rating prediction and content-based retrieval, among others.
\subsection{Experimental setup}

The MM-IMDb dataset has been split in three subsets. Train, development and test subsets contain $15552$, $2608$ and $7799$ respectively. The distribution of samples is listed in Table \ref{tab:Tags-distribution}. The sample was stratified so that training, dev and test sets comprises 60\%, 10\%, 30\% samples of each genre respectively.

In the multilabel classification the performance evaluation can be more complex than traditional \emph{multi-class} classification and the differences can be significant among several measures \citep{madjarov2012extensive}. Herein, four averages of the f-score ($f_1$) are reported: \emph{samples} computes the f-score per sample and then averages the results, \emph{micro} computes the f-score using all predictions at once, \emph{macro} computes the f-score per genre and then averages the results. \emph{weighted} is the same as \emph{macro} with a weighted average based on the number of positive samples per genre. Concretely, we calculate them as follows \citep{madjarov2012extensive}:
\begin{eqnarray*}
f_{1}^{sample} = \frac{1}{N}\sum_{i=1}^N \frac{2\times\left|\hat{y}_i\cap y_i\right|}{\left|\hat{y}_i\right| + \left|y_i\right|} & 
f_{1}^{macro} = \frac{1}{Q}\sum_{j=1}^Q \frac{2\times \textrm{p}_j \times \textrm{r}_j}{\textrm{p}_j + \textrm{r}_j} & 
f_{1}^{weighted} = \frac{1}{Q^2}\sum_{j=1}^Q Q_j\frac{2 \times \textrm{p}_j \times \textrm{r}_j}{\textrm{p}_j + \textrm{r}_j} 
\end{eqnarray*}
\begin{eqnarray*}
\text{p}^{micro} = \frac{\sum_{j=1}^Q tp_j}{\sum_{j=1}^Q tp_j + \sum_{j=1}^Q fp_j} & \text{r}^{micro} = \frac{\sum_{j=1}^Q tp_j}{\sum_{j=1}^Q tp_j + \sum_{j=1}^Q fn_j} &
f_{1}^{micro} = \frac{2\times \text{p}^{micro} \times \text{r}^{micro}}{\text{p}^{micro} + \text{r}^{micro}} \\
\end{eqnarray*}
With $N$ the number of examples; $Q$ the number of labels; $Q_j$ the number of true instances for the $j$-th label; $\textrm{p}$ the precision, $\textrm{r}$ the recall; $\hat{y}_i, y_i \in \left( 0,1 \right)^Q$ the prediction and ground truth binary tuples respectively; $tp_j, fp_j \textrm{and} fn_j$ the number of true positives, false positives and false negatives for the $j$-th label respectively.
\begin{table}
\centering{}
\caption{Genre distribution per subset}
\label{tab:Tags-distribution}
\begin{tabular}{llll|lllll}
\textbf{Genre} & \textbf{Train} & \textbf{Dev} & \textbf{Test} & & \textbf{Genre} & \textbf{Train} & \textbf{Dev} & \textbf{Test}
\\ \hline \\
Drama & 8424 & 1401 & 4142 & & Family & 978 & 172 & 518\\
Comedy & 5108 & 873 & 2611 & & Biography & 788 & 144 & 411\\
Romance & 3226 & 548 & 1590 & &  War & 806 & 128 & 401\\
Thriller & 3113 & 512 & 1567 & & History & 680 & 118 & 345\\
Crime & 2293 & 382 & 1163 & & Music & 634 & 100 & 311\\
Action &  2155 & 351 & 1044 & &  Animation & 586 & 105 & 306\\
Adventure & 1611 & 278 & 821 & &  Musical & 503 & 85 & 253\\
 Horror & 1603 & 275 & 825 & &  Western & 423 & 72 & 210\\
 Documentary & 1234 & 219 & 629 & &  Sport & 379 & 64 & 191\\
 Mystery & 1231 & 209 & 617 & &  Short & 281 & 48 & 142\\

 Sci-Fi & 1212 & 193 & 586 & &  Film-Noir & 202 & 34 & 102\\
 Fantasy & 1162 & 186 & 585 

\end{tabular}
\end{table}

\subsubsection*{Textual representation}
The pretrained Google Word2vec\footnote{\url{https://code.google.com/archive/p/word2vec/}} embedding space was used. After intersecting the Google word2vec available words with the MM-IMDb plots, the final vocabulary contains 41,612 words. Other than lowercase, no text preprocessing was applied. Since it is our intention to measure how the network's depth affects the performance of the model, we also evaluate the architecture with a single fully connected layer. In order to compare the performance of this textual representation, we evaluate it using two publicly available datasets: \emph{7genre} dataset that comprises 1,400 web pages with 7 disjoint genres and \emph{ki-04} dataset that comprises 1,239 samples classified under 8 genres. We compare the model with the state of the art results \citep{kanaris2009learning} which used character n-grams with structured information from the HTML tags to predict the genre of web pages.

\subsubsection*{Visual representation}
Since the first approach was to use VGG as a feature extractor. This model is referred as \emph{VGG\_Transfer}. The second approach takes as input the raw images to a CNN. Since all the images do not have the same size, all images were scaled, and cropped when required, to $160\times256$ pixels keeping the aspect ratio. This CNN comprises 5 CNN layers of $5,3,3,3,3$ squared filters and $2\times2$ pool sizes. Each convolutional layer has 16 hidden units. The convolutional layers are connected with the \emph{MaxoutMLP} on top. 

\subsubsection*{Multimodal representation}
We evaluate 4 different ways to combine both modalities as baselines.
\begin{description}
\item[{Average probability}] This can be seen as a late-fusion strategy. The probabilities obtained by the best model of each modality are averaged and thresholded.
\item[{concatenation}] Different works have found that a simple concatenation of representations of different modalities are good for combining the information \citep{suk2013deep,pei2013unsupervised,kiela2014learning}. Herein, we concatenated both representations to train the \emph{MaxoutMLP} architecture.
\item[{linear sum}] Following the way \citet{vinyals2015show} combine text and images representation into a single space, this model adds a linear transformation for each modality so that both outputs have the same size to be summed up and then followed by the \emph{MaxoutMLP} architecture.
\item[{MoE}] The mixture of experts (MoE) \citep{jacobs1991adaptive} model was adapted for multilabel classification. two gating strategies were explored: \emph{tied}, where a single gate multiplies all the logistics outputs, and \emph{untied} where every logistic output has its own gate. Logistic regression and \emph{MaxoutMLP} were evaluated as experts.
\end{description}

\subsubsection*{Neural network training}

Neural network models were trained using using Batch Normalization scheme \citep{ioffe2015batch}. This strategy applies a normalization step across samples that belong to the same batch, so that each hidden unit in the network receive a zero-mean and unit variance. Stochastic gradient descent with ADAM optimization \citep{kingma2014adam} was used to learn the weights of the neural network. Dropout and max-norm regularization were used to control overfitting. Hidden size (${\lbrace64, 128, 256, 512\rbrace}$), learning rate ($\left[10^{-3}, 10^{-1}\right]$), dropout ($\left[0.3, 0.7\right]$), max-norm ($\left[5, 20\right]$) and initialization ranges ($\left[10^{-3}, 10^{-1}\right]$) parameters were explored by training 25 models with random (uniform) hyperparameter initializations and the best was chosen according to validation performance. It has been reported that this strategy is preferable over grid search when training deep models \citep{bergstra2012random}. All the implementation was carried on with the Blocks framework \citep{van2015blocks}\footnote{\url{https://github.com/johnarevalo/gmu-mmimdb}}.

During the training process, we noticed that batch normalization considerably helped in terms of training time and convergence, resulting in less sensitivity to hyperparameters such as initialization ranges or learning rate. Also, dropout and max-norm regularization strategies helped to increase the performance at test time.

\section{Results}\label{sec:results}
\subsection{Evaluation over synthetic data}
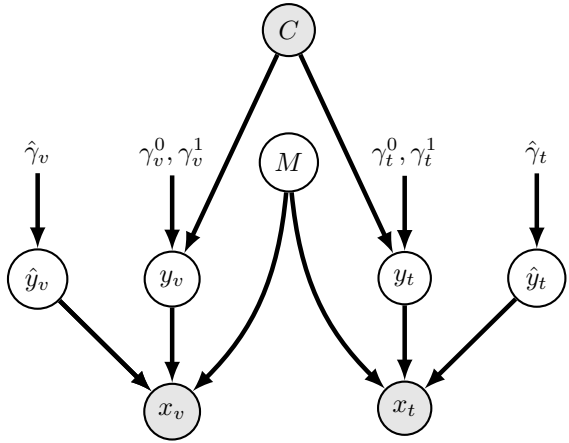
\begin{figure}
\centering{}
\begin{tikzpicture}[
cross/.style={path picture={
  \draw[black]
(path picture bounding box.south east) -- (path picture bounding box.north west) (path picture bounding box.south west) -- (path picture bounding box.north east);
}},
conn/.style={->, rounded corners, line width=0.60mm, -latex},
elemwise/.style={circle, draw, thick},
]
    \node[elemwise, fill=black!10] (c) {$C$};
    \node[elemwise, below=of c] (m) {$M$};
    \node[elemwise, below left=of m] (y_v) {$y_v$};
    \node[elemwise, below right=of m] (y_t) {$y_t$};
    \node[elemwise, below=of y_v, fill=black!10] (x_v) {$x_v$};
    \node[elemwise, left=of y_v] (y_hat_v) {$\hat{y}_v$};
    \node[above=of y_hat_v] (gamma_hat_v) {$\hat{\gamma}_v$};
    \node[above=of y_v] (gamma_v) {$\gamma_v^0,\gamma_v^1$};
    \node[above=of y_t] (gamma_t) {$\gamma_t^0,\gamma_t^1$};
    \node[elemwise, right=of y_t] (y_hat_t) {$\hat{y}_t$};
    \node[above=of y_hat_t] (gamma_hat_t) {$\hat{\gamma}_t$};

    \draw[conn] (gamma_v) -- (y_v);
    \draw[conn] (gamma_t) -- (y_t);
    \draw[conn] (gamma_hat_v) -- (y_hat_v);
    \draw[conn] (gamma_hat_t) -- (y_hat_t);
    \draw[conn] (y_hat_v) -- (x_v);

	\node[elemwise, below=of y_t, fill=black!10] (x_t) {$x_t$};

	\draw[conn] (c) -- (y_v);
    \draw[conn] (c) -- (y_t);
    \draw[conn] (m) to[bend right=20]  (x_t);
    \draw[conn] (m) to[bend left=20]  (x_v);
    \draw[conn] (y_t) -- (x_t);
    \draw[conn] (y_v) -- (x_v);
    \draw[conn] (y_hat_v) -- (x_v);
    \draw[conn] (y_hat_t) -- (x_t);
  \end{tikzpicture}
\caption{Generative model for the synthetic task. Grayed nodes represent visible variables, the other nodes represent hidden variables.}
\label{fig:synthetic}
\end{figure}

In order to evaluate if the model is able to identify which modality is contributing more information to classify a particular sample, we created a synthetic task based on a generative model, which is  depicted in Figure \ref{fig:synthetic}.
In this model we define the random binary variable $C$ as the target and $x_v, x_t \in \mathbb{R}^2$ as the input features. $M$ is a random binary variable that decides which modality will contain the relevant information that determines the class. The input features of each modality can be generated by a random source, $\hat{y}_v$ and $\hat{y}_t$, or by an informed source, $y_v$ and $y_t$. The generative model is specified as follows:

\noindent\begin{minipage}{.5\linewidth}
\centering{}
\begin{align*}
C & \sim\mathrm{Bernoulli}(p_{C})\\
M & \sim\mathrm{Bernoulli}(p_{M})\\
y_{v} & \sim\mathcal{N}(\gamma_{v}^{C})\\
\hat{y}_{v} & \sim\mathcal{N}(\hat{\gamma}_{v})\\
\end{align*}
\end{minipage}%
\begin{minipage}{.5\linewidth}
\centering{}
\begin{align*}
 x_{v} & =My_{v}+(1-M)\hat{y}_{v}\\
y_{t} & \sim\mathcal{N}(\gamma_{t}^{C})\\
\hat{y}_{t} & \sim\mathcal{N}(\hat{\gamma}_{t})\\
x_{t} & =M\hat{y}_{t}+(1-M)y_{t}\\
\end{align*}
\end{minipage}
We trained a model with a single GMU and applied a sigmoid function over $h$, then the binary cross entropy was used as loss function. Using the generative model, 200 samples per class were generated for each experiment. 1000 synthetic experiments with different random seeds were run and the GMU outperformed a logistic regression classifier in 370 of them, while obtaining equal results in the remainder ones. Our goal in these simulations was to show that the model was able to learn a latent variable that determines which modality carries the useful information for the classification. An interesting result is that between $M$ and the activations of the gate $z$ there is a correlation of $1$. This means the model was capable of learning such latent variable by only observing the $x_v$ and $x_t$ input features.

\begin{figure}
\begin{minipage}{.5\textwidth}
\centering{}\includegraphics[width=\textwidth]{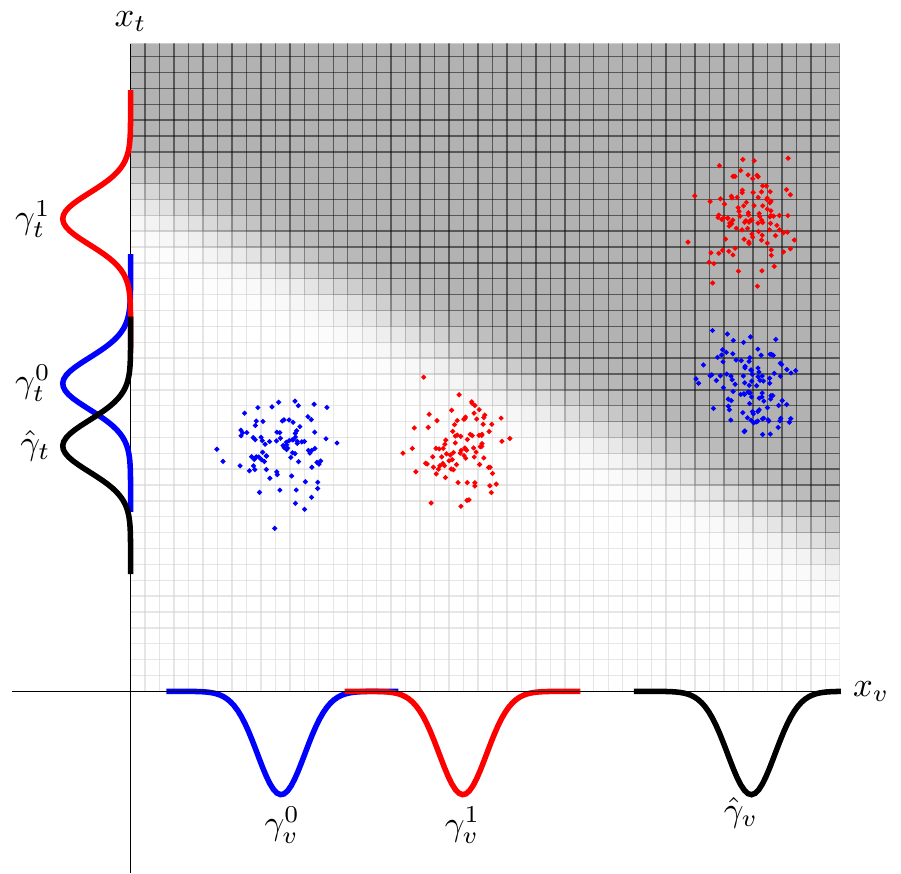}
\end{minipage}
\begin{minipage}{.5\textwidth}
\centering{}\includegraphics[width=\textwidth]{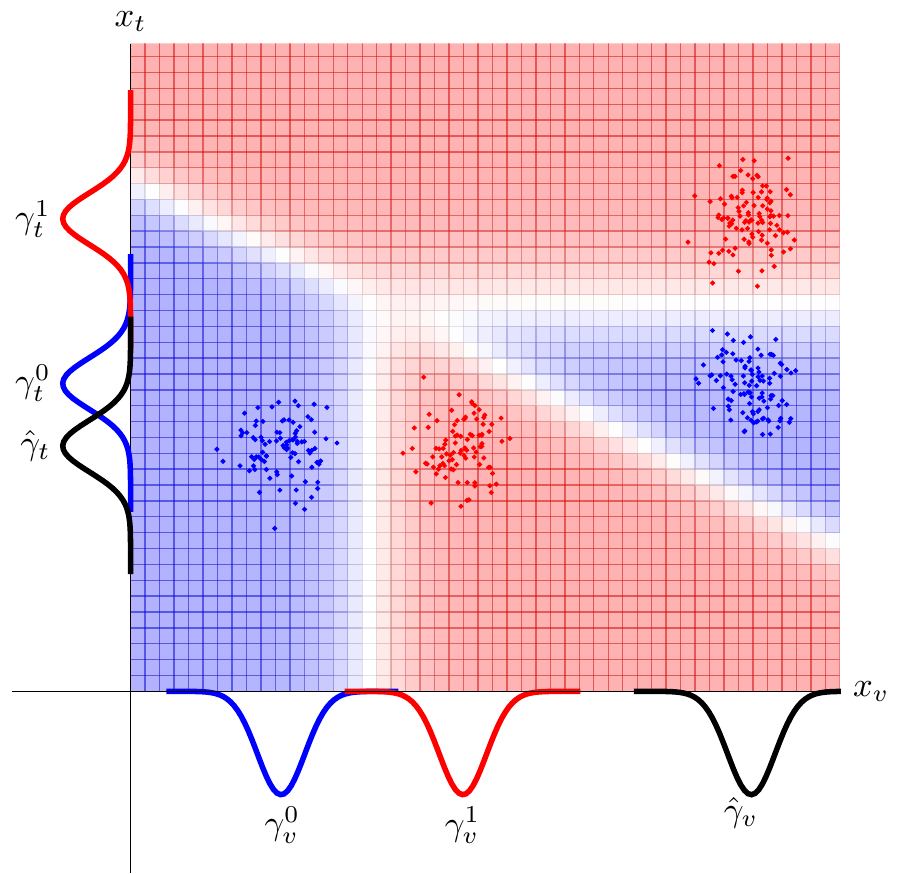}
\end{minipage}
\caption{Activations of $z$ (left) and prediction (right) for a synthetic experiment with $x_v,x_t\in \mathbb{R}$. \label{fig:zacts}}
\end{figure}

We also wanted to project back the $z$ activations to the feature space in order to visualize regions depending on the modality. Figure \ref{fig:zacts} shows the activations in a synthetic experiment generated by the setup of Figure \ref{fig:synthetic} for $x_v, x_t \in \mathbb{R}$. Each axis represents a modality, red and blue dots are the samples generated for the two classes and black Gaussian curves represent the $\hat{\gamma}_v$ and $\hat{\gamma}_t$ noises. The contour of the left figure (gray) represents the activation of $z$. Notice that in white regions ($z=1$), the model gives more importance to the $x_v$ modality while in gray regions ($z=0$) the $x_t$ modality is more relevant; i.e. the $z$ gate is isolating the noise. The contour of the right figure (blue-red) represents the model prediction. It is noteworthy that the boundary defined by the gates still holds when the model solves the task. This also encourages the inclusion of non-linearities to the $z$ gate so that it is able to discriminate more complex interactions between modalities.

\subsection{Genre classification results}
Before using our text representation in the multimodal task, we wanted to be sure such representation was good enough to address the genre classification task. Thus, we evaluated it on 2 public datasets. We found \emph{MaxoutMLP\_w2v} achieves the state of the art results on the \emph{ki-04} dataset and increases the performance in the \emph{7Genre} dataset from $0.841$ to $0.854$ \citep{kanaris2009learning}. Notice that the baseline uses additional information from the HTML structure from the web page, while this representation uses only the text data.

\begin{table}[!ht]
\centering
\caption{Summary of classification task on the MM-IMDb dataset}
\label{tab:mmimdb-results}
\begin{tabular}{llllll}
\multirow{2}{*}{\textbf{Modality}} & \multirow{2}{*}{\textbf{Representation}} & \multicolumn{4}{c}{\textbf{F-Score}}         \\
                          &                                 & weighted & samples & micro & macro \\
\hline \\
Multimodal                & GMU                             & \textbf{0.617}    & \textbf{0.630}   & \textbf{0.630} & \textbf{0.541} \\
                          & Linear\_sum         & 0.600    & 0.607   & 0.607 & 0.530 \\
                          & Concatenate         & 0.597    & 0.605   & 0.606 & 0.521 \\
                          & AVG\_probs                      & 0.604    & 0.616   & 0.615 & 0.491 \\
                          & MoE\_MaxoutMLP & 0.592 & 0.593 & 0.601 & 0.516\\
                          & MoE\_MaxoutMLP (tied) & 0.579 & 0.579 & 0.587 & 0.489\\
                          & MoE\_Logistic & 0.541 & 0.557 & 0.565 & 0.456\\
                          & MoE\_Logistic (tied) & 0.483 & 0.507 & 0.518 & 0.358\\
\hline \\
\multirow{7}{*}{Text}     & MaxoutMLP\_w2v                  & 0.588    & 0.592   & 0.595 & 0.488 \\
                          & RNN\_transfer                   & 0.570    & 0.580   & 0.580 & 0.480 \\
                          & MaxoutMLP\_w2v\_1\_hidden       & 0.540    & 0.540   & 0.550 & 0.440 \\
                          & Logistic\_w2v                   & 0.530    & 0.540   & 0.550 & 0.420 \\
                          & MaxoutMLP\_3grams             & 0.510    & 0.510   & 0.520 & 0.420 \\
                          & Logistic\_3grams                & 0.510    & 0.520   & 0.530 & 0.400 \\
                          & RNN\_end2end                    & 0.490    & 0.490   & 0.490 & 0.370 \\
\hline \\
\multirow{2}{*}{Visual}   & VGG\_Transfer                   & 0.410    & 0.429   & 0.437 & 0.284 \\
                          & CNN\_end2end                    & 0.370    & 0.350   & 0.340 & 0.210
\end{tabular}
\end{table}
Table \ref{tab:mmimdb-results} shows the results in the proposed dataset. For the textual modality, the best performance is obtained by the combination of word2vec representation with an MLP classifier. The behavior of all representation methods are consistent across the performance measures. Learning from scratch the RNN model performed the worst. We hypothesize this has to do with the lack of data to learn meaningful relations among words. It has been shown that millions of words are required to train a model such as word2vec that is able to exploit common regularities between word co-occurrences.

\begin{table}
\centering
\caption{Macro F-Score reported per genre for single and multimodal approaches.}
\label{tab:genre_results}
\begin{tabular}{llll|llll}
\textbf{Genre}       & \textbf{Textual} & \textbf{Visual} & \textbf{GMU}  &\textbf{Genre}       & \textbf{Textual} & \textbf{Visual} & \textbf{GMU}  \\
Drama       & 0.74    & 0.67   & \textbf{0.77} &Fantasy     & 0.42    & 0.25   & \textbf{0.46} \\
Comedy      & 0.65    & 0.59   & \textbf{0.68} &Family      & 0.5     & 0.46   & \textbf{0.58} \\
Romance     & \textbf{0.53}    & 0.33   & 0.51 &Biography   & \textbf{0.4}     & 0.02   & 0.25 \\
Thriller    & 0.57    & 0.39   & \textbf{0.62} &War         & 0.57    & 0.19   & \textbf{0.64} \\
Crime       & \textbf{0.61}    & 0.25   & 0.59 &History     & \textbf{0.35}    & 0.06   & 0.29 \\
Action      & 0.58    & 0.37   & \textbf{0.6}  &Animation   & 0.43    & 0.61   & \textbf{0.68} \\
Adventure   & \textbf{0.51}    & 0.32   & \textbf{0.51} &Musical     & 0.14    & 0.18   & \textbf{0.28} \\
Horror      & 0.65    & 0.41   & \textbf{0.69} &Western     & 0.52    & 0.37   & \textbf{0.65} \\
Documentary & 0.67    & 0.18   & \textbf{0.76} &Sport       & 0.64    & 0.11   & \textbf{0.7}  \\
Mystery     & 0.38    & 0.11   & \textbf{0.39} &Short       & 0.2     & 0.24   & \textbf{0.27} \\
Sci-Fi      & 0.63    & 0.3    & \textbf{0.66} &Film-Noir   & 0.02    & 0.11   & \textbf{0.37} \\
Music       & \textbf{0.51}    & 0.01    & 0.48 & & & & \\
\end{tabular}
\end{table}

For the visual modality, the usage of pretrained models works better than training the model from scratch. It seems it is still a small dataset to learn all the complexities of the posters. Now, comparing the performance independently per genre, as in Table \ref{tab:genre_results}, it is interesting to notice that in \emph{Animation} the visual modality outperforms the textual one.

In the multimodal scenario, by adding the GMU as building block to learn the fusion we obtained the best performance, improving independent modalities in the averaged measures and in 16 of out 23 genres and outperforming all other evaluated fusion strategies. The concatenation or the linear combination approaches were not enough to model the correlation between the modalities and MoE models did not perform better than simpler approaches. This is an expected behavior for MoE in a relatively small dataset because the data is fractionated over different experts, and thus it doesn't make an efficient use of the training samples.

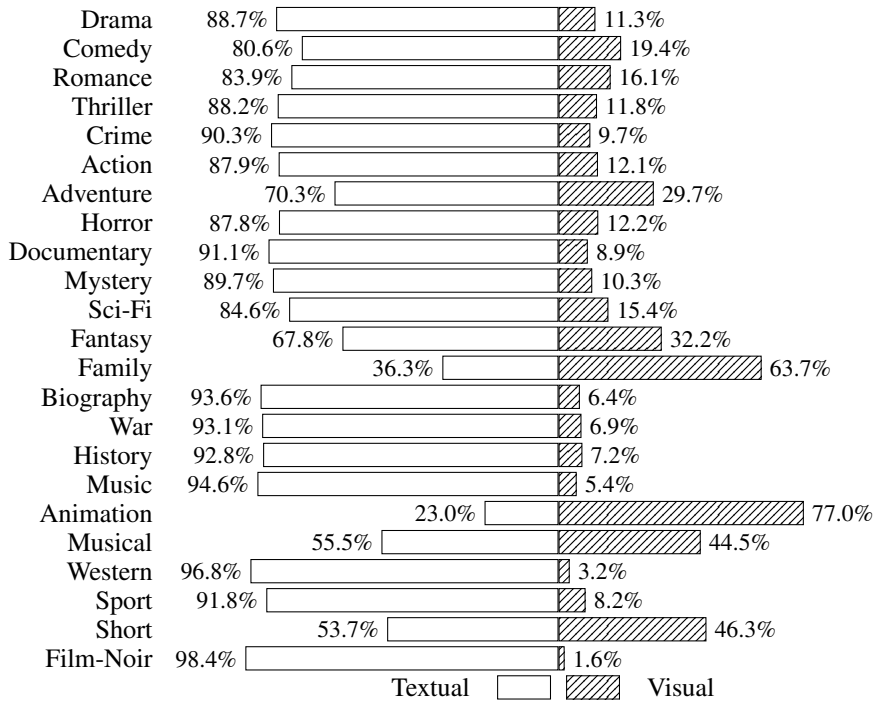
\begin{figure}
\centering{}
\pgfplotstableread[col sep=comma]{
age,man,woman
Drama,88.7,11.3
Comedy,80.6,19.4
Romance,83.9,16.1
Thriller,88.2,11.8
Crime,90.3,9.7
Action,87.9,12.1
Adventure,70.3,29.7
Horror,87.8,12.2
Documentary,91.1,8.9
Mystery,89.7,10.3
Sci-Fi,84.6,15.4
Fantasy,67.8,32.2
Family,36.3,63.7
Biography,93.6,6.4
War,93.1,6.9
History,92.8,7.2
Music,94.6,5.4
Animation,23.0,77.0
Musical,55.5,44.5
Western,96.8,3.2
Sport,91.8,8.2
Short,53.7,46.3
Film-Noir,98.4,1.6
}\loadedtable

\newlength{\dy}\setlength{\dy}{\baselineskip}
\newlength{\dx}\setlength{\dx}{0.12em}
\newlength{\temp}

\pgfplotstablegetrowsof{\loadedtable}
\pgfmathparse{\pgfplotsretval-1}
\edef\rows{\pgfmathresult}
\setlength{\temp}{0pt}
\foreach \y in {0,1,...,\rows}{%
  \pgfplotstablegetelem{\y}{man}\of\loadedtable
  \global\advance\temp by \pgfplotsretval pt
  \pgfplotstablegetelem{\y}{woman}\of\loadedtable
  \global\advance\temp by \pgfplotsretval pt
}
\pgfmathparse{0.01\temp}
\edef\total{\pgfmathresult}

\noindent\begin{tikzpicture}

\node at (0,2\dy) {};

\foreach \y in {0,1,...,\rows}{%
  \pgfplotstablegetelem{\y}{age}\of\loadedtable
  \node[left] at (-15em,-\y\dy) {\strut\pgfplotsretval};
  \pgfplotstablegetelem{\y}{man}\of\loadedtable
  \node[left,draw,text width=\pgfplotsretval\dx,text height=.8\dy,inner sep=0]
    at (0,-\y\dy+0.1\dy) {};
  \pgfplotstablegetelem{\y}{man}\of\loadedtable
  \node[left] at (-\pgfplotsretval\dx,-\y\dy) {\small\strut \pgfplotsretval\%};
  \pgfplotstablegetelem{\y}{woman}\of\loadedtable
  \node[left] at (17em,-\y\dy) {};
  \node[draw,right,pattern=north east lines,text width=\pgfplotsretval\dx,text height=.8\dy,inner sep=0]
    at (0,-\y\dy+0.1\dy) {};
  \pgfmathparse{\pgfplotsretval/\total}
  \pgfmathprintnumberto[fixed,precision=2]{\pgfmathresult}{\per}
  \pgfplotstablegetelem{\y}{woman}\of\loadedtable
  \node[right] at (\pgfplotsretval\dx,-\y\dy) {\small\strut\pgfplotsretval\%};
}
\node[left] at (-3em,-\rows\dy-.9\dy) {Textual};
\node[left,draw,text width=2em,text height=.8\dy,inner sep=0]
  at (-1mm,-\rows\dy-.9\dy) {};
\node[right,pattern=north east lines,draw,text width=2em,text height=.8\dy,inner sep=0]
  at (+1mm,-\rows\dy-.9\dy) {};
\node[right] at (3em,-\rows\dy-.9\dy) {Visual};

\end{tikzpicture}
\caption{Percentage of gates activations ($z>0.5$: Visual; $z<=0.5$: textual) for test samples to which the model assigned them the label.}
\label{fig:tornado}
\end{figure}
In order to evaluate which modality influences more the model when assigning a particular label, we averaged the activations of a subset of $z$ gates of the test samples to which the model assigned them such label. We counted the number of samples that pays more attention to the textual modality ($z<=0.5$) or to the visual modality ($z>0.5$). The units were chosen taking into account the mutual information between the predictions and the $z$ activations. The result of this analysis is depicted in Figure \ref{fig:tornado}. As expected, the model is generally more influenced by the textual modality. But, in particular cases such as \emph{Animation} or \emph{Family} genres, the visual modality affects more the model. This is also consistent with results of Table \ref{tab:genre_results} which reports better performances for visual modality.

\begin{table}
\caption{Examples of predictions in test set. Red and blue genres are false positives and true positives respectively.}
\label{tab:examples}
\centering{}
\begin{tabular}{|c|l|l|}
\hline 
\multicolumn{3}{|c|}{The World According to Sesame Street}\tabularnewline
\hline 
\multirow{5}{*}{\includegraphics[width=0.14\textwidth]{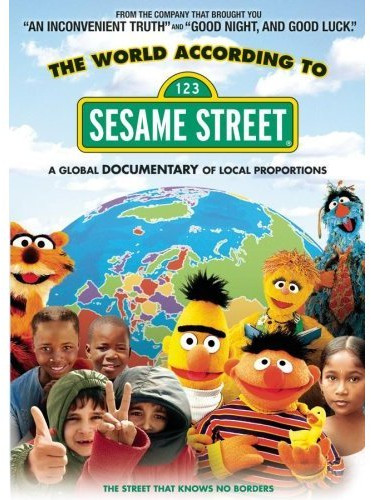}} & \multicolumn{2}{p{0.7\textwidth}|}{a documentary which examines the creation and co - production of the popular children ' s television program in three developing countries: bangladesh , kosovo and south africa .}\tabularnewline
\cline{2-3} 
 & Ground Truth & Documentary\tabularnewline
\cline{2-3} 
 & Textual & \textcolor{blue}{Documentary}, \textcolor{red}{History}\tabularnewline
\cline{2-3} 
 & Visual & \textcolor{blue}{Comedy}, \textcolor{red}{Adventure}, \textcolor{red}{Family}, \textcolor{red}{Animation}\tabularnewline
\cline{2-3} 
 & GMU & \textcolor{blue}{Documentary}\tabularnewline
\hline 
\end{tabular}
\begin{tabular}{|l|l|l|}
\hline 
\multicolumn{3}{|c|}{Babar: the movie}\tabularnewline
\hline 
  \multirow{5}{*}{\includegraphics[width=0.14\textwidth]{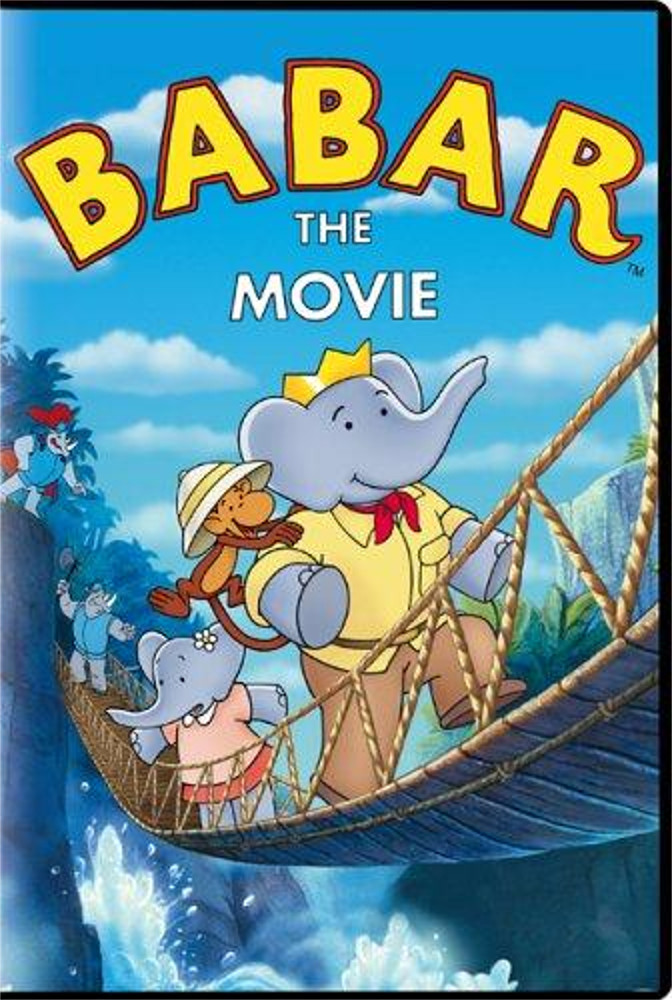}} & \multicolumn{2}{p{0.7\textwidth}|}{in his spectacular film debut , young babar , king of the elephants , must save his homeland from certain destruction by rataxes and his band of invading rhinos .  \newline}\tabularnewline
\cline{2-3} 
 & Ground Truth & Adventure, Fantasy, Family, Animation, Musical\tabularnewline
\cline{2-3} 
 & Textual & \textcolor{blue}{Adventure}, \textcolor{red}{Documentary}, \textcolor{red}{War}, \textcolor{red}{Music}\tabularnewline
\cline{2-3} 
 & Visual & \textcolor{red}{Comedy}, \textcolor{blue}{Adventure}, \textcolor{blue}{Family}, \textcolor{blue}{Animation}\tabularnewline
\cline{2-3} 
 & GMU & \textcolor{blue}{Adventure}, \textcolor{blue}{Family}, \textcolor{blue}{Animation}\tabularnewline
\hline 
\end{tabular}
\begin{tabular}{|l|l|l|}
\hline 
\multicolumn{3}{|c|}{Letters from Iwo Jima}\tabularnewline
\hline 
\multirow{5}{*}{\includegraphics[width=0.14\textwidth]{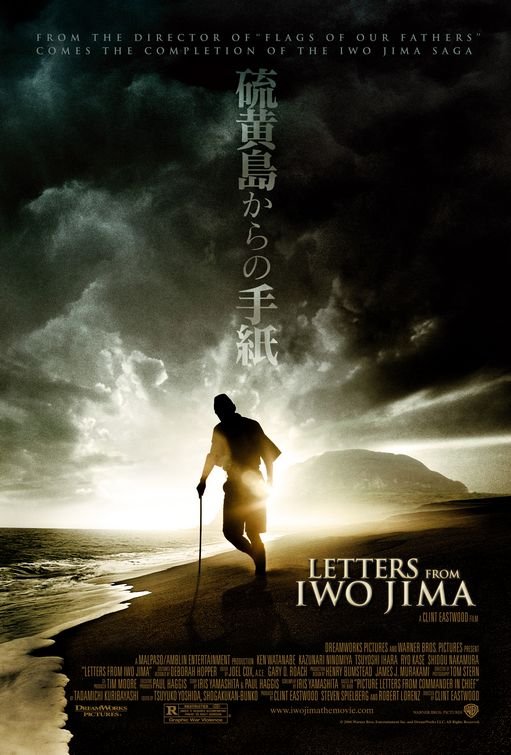}} & \multicolumn{2}{p{0.7\textwidth}|}{the island of iwo jima stands between the american military force and the home islands of japan . (...) when the american invasion begins , both kuribayashi and saigo find strength , honor , courage , and horrors beyond imagination .}\tabularnewline
\cline{2-3} 
 & Ground Truth & Drama, War, History\tabularnewline
\cline{2-3} 
 & Textual & \textcolor{blue}{Drama}, \textcolor{red}{Action}, \textcolor{blue}{War}, \textcolor{blue}{History}\tabularnewline
\cline{2-3} 
 & Visual & \textcolor{red}{Thriller}, \textcolor{red}{Action}, \textcolor{red}{Adventure}, \textcolor{red}{Sci-Fi}\tabularnewline
\cline{2-3} 
 & GMU & \textcolor{blue}{Drama}, \textcolor{blue}{War}, \textcolor{blue}{History}\tabularnewline
\hline 
\end{tabular}
\begin{tabular}{|l|l|l|}
\hline 
\multicolumn{3}{|c|}{The Last Elvis}\tabularnewline
\hline 
\multirow{5}{*}{\includegraphics[width=0.14\textwidth]{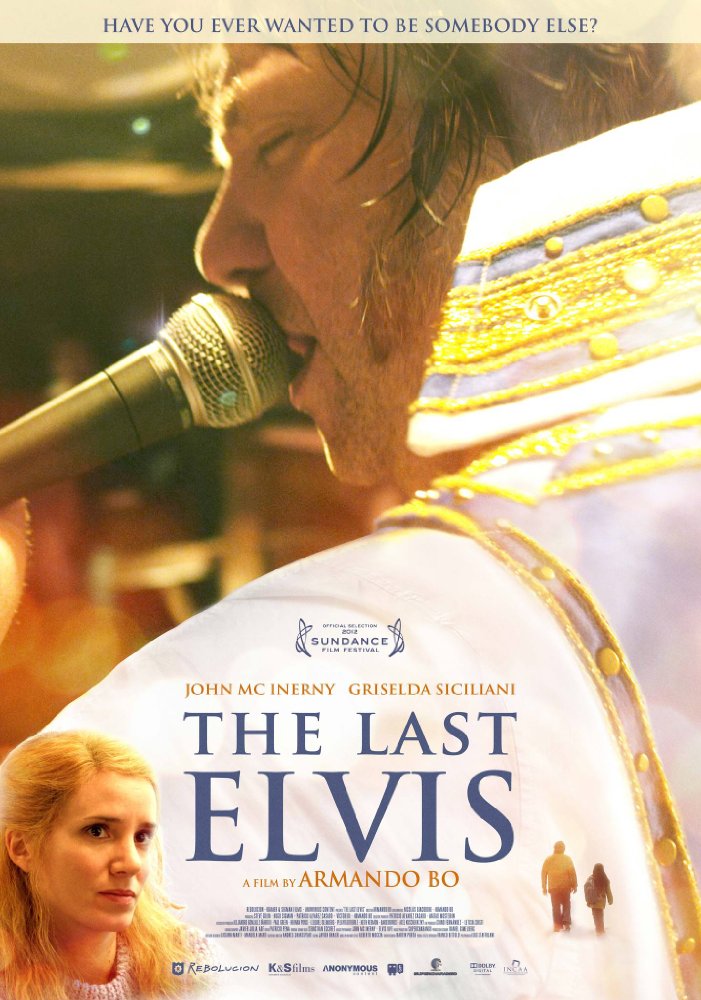}} & \multicolumn{2}{p{0.7\textwidth}|}{ a tragic accident causes an elvis impersonator to reassess his priorities. \newline \newline }\tabularnewline
\cline{2-3} 
 & Ground Truth & Drama\tabularnewline
\cline{2-3} 
 & Textual & \textcolor{red}{Comedy}, \textcolor{red}{Documentary}, \textcolor{red}{Family}, \textcolor{red}{Biography}, \textcolor{red}{Music}\tabularnewline
\cline{2-3} 
 & Visual & \textcolor{blue}{Drama}, \textcolor{red}{Romance} \tabularnewline
\cline{2-3} 
 & GMU & \textcolor{blue}{Drama} \tabularnewline
\hline 
\end{tabular}
\end{table}

We wanted to qualitative explore test examples in which performance was improved by a relative large margin. Table \ref{tab:examples} illustrates cases where the model takes advantage of the most accurate modality, and in some cases removes false positives. It is noteworthy that some of these examples can be confusing for a human if one modality is missing, or additional context information is not given.

\section{Conclusions}\label{sec:conclusions}

This work presented a strategy to learn fusion transformations from multimodal sources. Similarly to the way recurrent models control the information flow, the proposed model is based on multiplicative gates. The Gated Multimodal Unit (GMU) receives two or more input sources and learns to determine how much each input modality affects the unit activation. In synthetic experiments the GMU was able to learn hidden latent variables, and in a real scenario it outperformed the single-modality approaches. An interesting property of GMU is that, being a differentiable operation, it is easily coupled in any other neural network architecture and trained with standard gradient-based optimization algorithms. With this work we will also release a new dataset that contains around $27,000$ movie plots, images and other metadata. To the best of our knowledge, this is the biggest dataset used to perform movie genre classification based on multimodal information and the first one to be publicly available. In our future work we expect to explore deep architectures of GMU layers as well as integration with attention mechanism over the input modalities. Also, It will be interesting to explore in more depth the interpretability of the learned features.

\subsubsection*{Acknowledgments}

Arevalo thanks Colciencias for its support through a doctoral grant in call 617/2013. The authors also thank for K40 Tesla GPU donated by NVIDIA and which was used for some representation learning experiments.

\bibliography{iclr2017_conference}

\begin{thebibliography}{54}
\providecommand{\natexlab}[1]{#1}
\providecommand{\url}[1]{\texttt{#1}}
\expandafter\ifx\csname urlstyle\endcsname\relax
  \providecommand{\doi}[1]{doi: #1}\else
  \providecommand{\doi}{doi: \begingroup \urlstyle{rm}\Url}\fi

\bibitem[Akata et~al.(2014)Akata, Lee, and Schiele]{akata2014zero}
Zeynep Akata, Honglak Lee, and Bernt Schiele.
\newblock {Zero-Shot Learning with Structured Embeddings}.
\newblock \emph{CoRR}, abs/1409.8, 2014.
\newblock URL \url{http://arxiv.org/abs/1409.8403}.

\bibitem[Anand(2014)]{anand2014evaluating}
Deepa Anand.
\newblock {Evaluating folksonomy information sources for genre prediction}.
\newblock In \emph{Advance Computing Conference (IACC), 2014 IEEE
  International}, pp.\  887--892, feb 2014.
\newblock \doi{10.1109/IAdCC.2014.6779440}.

\bibitem[Andrew et~al.(2013)Andrew, Arora, Bilmes, and Livescu]{andrew2013deep}
Galen Andrew, Raman Arora, Jeff~A Bilmes, and Karen Livescu.
\newblock Deep canonical correlation analysis.
\newblock In \emph{ICML (3)}, pp.\  1247--1255, 2013.

\bibitem[Antol et~al.(2015)Antol, Agrawal, Lu, Mitchell, Batra, Zitnick, and
  Parikh]{antol2015vqa}
Stanislaw Antol, Aishwarya Agrawal, Jiasen Lu, Margaret Mitchell, Dhruv Batra,
  C.~Lawrence Zitnick, and Devi Parikh.
\newblock Vqa: Visual question answering.
\newblock In \emph{International Conference on Computer Vision (ICCV)}, 2015.

\bibitem[Atrey et~al.(2010)Atrey, Hossain, {El Saddik}, and
  Kankanhalli]{atrey2010multimodal}
Pradeep~K. Atrey, M.~Anwar Hossain, Abdulmotaleb {El Saddik}, and Mohan~S.
  Kankanhalli.
\newblock {Multimodal fusion for multimedia analysis: a survey}.
\newblock \emph{Multimedia Systems}, 16\penalty0 (6):\penalty0 345--379, April
  2010.
\newblock ISSN 0942-4962.
\newblock \doi{10.1007/s00530-010-0182-0}.

\bibitem[Bengio et~al.(2003)Bengio, Ducharme, Vincent, and
  Janvin]{bengio2003neural}
Yoshua Bengio, R{\'e}jean Ducharme, Pascal Vincent, and Christian Janvin.
\newblock A neural probabilistic language model.
\newblock \emph{The Journal of Machine Learning Research}, 3:\penalty0
  1137--1155, 2003.

\bibitem[Bergstra \& Bengio(2012)Bergstra and Bengio]{bergstra2012random}
James Bergstra and Yoshua Bengio.
\newblock Random search for hyper-parameter optimization.
\newblock \emph{Journal of Machine Learning Research}, 13\penalty0
  (Feb):\penalty0 281--305, 2012.

\bibitem[Bhatt \& Kankanhalli(2011)Bhatt and Kankanhalli]{bhatt2011multimedia}
Chidansh Bhatt and Mohan Kankanhalli.
\newblock {Multimedia data mining: state of the art and challenges}.
\newblock \emph{Multimedia Tools and Applications}, 51\penalty0 (1):\penalty0
  35--76, 2011.
\newblock ISSN 1380-7501.
\newblock \doi{10.1007/s11042-010-0645-5}.

\bibitem[Coates \& Ng(2011)Coates and Ng]{coates2011importance}
Adam Coates and Andrew~Y Ng.
\newblock The importance of encoding versus training with sparse coding and
  vector quantization.
\newblock In \emph{Proceedings of the 28th International Conference on Machine
  Learning (ICML-11)}, pp.\  921--928, 2011.

\bibitem[Deng(2014)]{deng2014tutorial}
Li~Deng.
\newblock {A tutorial survey of architectures, algorithms, and applications for
  deep learning}.
\newblock \emph{APSIPA Transactions on Signal and Information Processing}, 3,
  2014.
\newblock ISSN 2048-7703.
\newblock \doi{10.1017/atsip.2013.9}.
\newblock URL \url{http://journals.cambridge.org/article{\_}S2048770313000097}.

\bibitem[Feng et~al.(2013)Feng, Li, and Wang]{feng2013constructing}
Fangxiang Feng, Ruifan Li, and Xiaojie Wang.
\newblock Constructing hierarchical image-tags bimodal representations for word
  tags alternative choice.
\newblock \emph{arXiv preprint arXiv:1307.1275}, 2013.

\bibitem[Frome et~al.(2013)Frome, Corrado, Shlens, Bengio, Dean, Ranzato, and
  Mikolov]{frome2013devise}
Andrea Frome, Greg~S Corrado, Jon Shlens, Samy Bengio, Jeff Dean,
  Marc$\backslash$textquotesingle~Aurelio Ranzato, and Tomas Mikolov.
\newblock {DeViSE: A Deep Visual-Semantic Embedding Model}.
\newblock In C~J~C Burges, L~Bottou, M~Welling, Z~Ghahramani, and K~Q
  Weinberger (eds.), \emph{Advances in Neural Information Processing Systems
  26}, pp.\  2121--2129. Curran Associates, Inc., 2013.
\newblock URL
  \url{http://papers.nips.cc/paper/5204-devise-a-deep-visual-semantic-embedding-model.pdf}.

\bibitem[Fu et~al.(2015)Fu, Li, Li, and Wei]{fu2015fast}
Zhikang Fu, Bing Li, Jun Li, and Shuhua Wei.
\newblock {Fast Film Genres Classification Combining Poster and Synopsis}.
\newblock In Xiaofei He, Xinbo Gao, Yanning Zhang, Zhi-Hua Zhou, Zhi-Yong Liu,
  Baochuan Fu, Fuyuan Hu, and Zhancheng Zhang (eds.), \emph{Lecture Notes in
  Computer Science}, volume 9242 of \emph{Lecture Notes in Computer Science},
  pp.\  72--81. Springer International Publishing, Cham, 2015.
\newblock \doi{10.1007/978-3-319-23989-7_8}.
\newblock URL \url{http://link.springer.com/10.1007/978-3-319-23862-3
  http://link.springer.com/10.1007/978-3-319-23989-7{\_}8}.

\bibitem[Goodfellow et~al.(2013)Goodfellow, Warde-farley, Mirza, Courville, and
  Bengio]{goodfellow2013maxout}
Ian Goodfellow, David Warde-farley, Mehdi Mirza, Aaron Courville, and Yoshua
  Bengio.
\newblock Maxout networks.
\newblock In Sanjoy Dasgupta and David Mcallester (eds.), \emph{Proceedings of
  the 30th International Conference on Machine Learning (ICML-13)}, volume~28,
  pp.\  1319--1327. JMLR Workshop and Conference Proceedings, May 2013.

\bibitem[Hong \& Hwang(2015)Hong and Hwang]{hong2015multimodal}
Hao-Zhi Hong and Jen-Ing~G Hwang.
\newblock {Multimodal PLSA for Movie Genre Classification}.
\newblock In Friedhelm Schwenker, Fabio Roli, and Josef Kittler (eds.),
  \emph{Multiple Classifier Systems: 12th International Workshop, MCS 2015,
  G{\{}{\"{u}}{\}}nzburg, Germany, June 29 - July 1, 2015, Proceedings}, pp.\
  159--167. Springer International Publishing, Cham, 2015.
\newblock ISBN 978-3-319-20248-8.
\newblock \doi{10.1007/978-3-319-20248-8_14}.
\newblock URL \url{http://dx.doi.org/10.1007/978-3-319-20248-8{\_}14}.

\bibitem[Huang et~al.(2012)Huang, Socher, Manning, and Ng]{huang2012improving}
Eric~H Huang, Richard Socher, Christopher~D Manning, and Andrew Ng.
\newblock {Improving word representations via global context and multiple word
  prototypes}.
\newblock In \emph{Proceedings of the 50th Annual Meeting of the Association
  for Computational Linguistics: Long Papers-Volume 1}, pp.\  873--882.
  Association for Computational Linguistics, 2012.

\bibitem[Huang et~al.(2007)Huang, Shih, and Hsu]{huang2007film}
Hui-Yu Huang, Weir-Sheng Shih, and Wen-Hsing Hsu.
\newblock {A Film Classifier Based on Low-level Visual Features}.
\newblock In \emph{2007 IEEE 9th Workshop on Multimedia Signal Processing},
  volume~3, pp.\  465--468. IEEE, 2007.
\newblock ISBN 978-1-4244-1273-0.
\newblock \doi{10.1109/MMSP.2007.4412917}.
\newblock URL
  \url{http://ieeexplore.ieee.org/lpdocs/epic03/wrapper.htm?arnumber=4412917}.

\bibitem[Ioffe \& Szegedy(2015)Ioffe and Szegedy]{ioffe2015batch}
Sergey Ioffe and Christian Szegedy.
\newblock Batch normalization: Accelerating deep network training by reducing
  internal covariate shift.
\newblock In \emph{Proceedings of The 32nd International Conference on Machine
  Learning}, pp.\  448--456, 2015.

\bibitem[Ivasic-Kos et~al.(2014)Ivasic-Kos, Pobar, and
  Mikec]{ivasic-kos2014movie}
Marina Ivasic-Kos, Miran Pobar, and Luka Mikec.
\newblock {Movie posters classification into genres based on low-level
  features}.
\newblock In \emph{2014 37th International Convention on Information and
  Communication Technology, Electronics and Microelectronics (MIPRO)},
  volume~i, pp.\  1198--1203. IEEE, may 2014.
\newblock ISBN 978-953-233-077-9.
\newblock \doi{10.1109/MIPRO.2014.6859750}.
\newblock URL
  \url{http://ieeexplore.ieee.org/lpdocs/epic03/wrapper.htm?arnumber=6859750}.

\bibitem[Ivasic-Kos et~al.(2015)Ivasic-Kos, Pobar, and
  Ipsic]{ivasic-kos2015automatic}
Marina Ivasic-Kos, Miran Pobar, and Ivo Ipsic.
\newblock {Automatic Movie Posters Classification into Genres}.
\newblock In Madevska~Ana Bogdanova and Dejan Gjorgjevikj (eds.), \emph{ICT
  Innovations 2014: World of Data}, pp.\  319--328. Springer International
  Publishing, Cham, 2015.
\newblock ISBN 978-3-319-09879-1.
\newblock \doi{10.1007/978-3-319-09879-1_32}.
\newblock URL \url{http://dx.doi.org/10.1007/978-3-319-09879-1{\_}32}.

\bibitem[Jacobs et~al.(1991)Jacobs, Jordan, Nowlan, and
  Hinton]{jacobs1991adaptive}
Robert~A Jacobs, Michael~I Jordan, Steven~J Nowlan, and Geoffrey~E Hinton.
\newblock Adaptive mixtures of local experts.
\newblock \emph{Neural computation}, 3\penalty0 (1):\penalty0 79--87, 1991.

\bibitem[Johnson et~al.(2015)Johnson, Karpathy, and
  Fei-Fei]{johnson2015densecap}
Justin Johnson, Andrej Karpathy, and Li~Fei-Fei.
\newblock Densecap: Fully convolutional localization networks for dense
  captioning.
\newblock \emph{arXiv preprint arXiv:1511.07571}, 2015.

\bibitem[Kanaris \& Stamatatos(2009)Kanaris and
  Stamatatos]{kanaris2009learning}
Ioannis Kanaris and Efstathios Stamatatos.
\newblock {Learning to recognize webpage genres}.
\newblock \emph{Information Processing and Management}, 45\penalty0
  (5):\penalty0 499--512, 2009.
\newblock ISSN 03064573.
\newblock \doi{10.1016/j.ipm.2009.05.003}.
\newblock URL \url{http://dx.doi.org/10.1016/j.ipm.2009.05.003}.

\bibitem[Kang et~al.(2012)Kang, Kim, and Choi]{kang2012deep}
Yoonseop Kang, Saehoon Kim, and Seungjin Choi.
\newblock Deep learning to hash with multiple representations.
\newblock In \emph{2012 IEEE 12th International Conference on Data Mining},
  pp.\  930--935. IEEE, 2012.

\bibitem[Kiela \& Bottou(2014)Kiela and Bottou]{kiela2014learning}
Douwe Kiela and L\'{e}on Bottou.
\newblock {Learning Image Embeddings using Convolutional Neural Networks for
  Improved Multi-Modal Semantics}.
\newblock In \emph{{Proceedings of the Conference on Empirical Methods in
  Natural Language Processing (EMNLP-14)}}, 2014.

\bibitem[Kingma \& Ba(2014)Kingma and Ba]{kingma2014adam}
Diederik Kingma and Jimmy Ba.
\newblock Adam: A method for stochastic optimization.
\newblock \emph{arXiv preprint arXiv:1412.6980}, 2014.

\bibitem[Kiros et~al.(2014{\natexlab{a}})Kiros, Salakhutdinov, and
  Zemel]{kiros2013multimodal}
Ryan Kiros, Ruslan Salakhutdinov, and Richard~S Zemel.
\newblock Multimodal neural language models.
\newblock In \emph{ICML}, volume~14, pp.\  595--603, 2014{\natexlab{a}}.

\bibitem[Kiros et~al.(2014{\natexlab{b}})Kiros, Salakhutdinov, and
  Zemel]{kiros2014unifying}
Ryan Kiros, Ruslan Salakhutdinov, and Richard~S Zemel.
\newblock {Unifying Visual-Semantic Embeddings with Multimodal Neural Language
  Models}.
\newblock \emph{arXiv preprint arXiv:1411.2539}, 2014{\natexlab{b}}.

\bibitem[LeCun et~al.(2015)LeCun, Bengio, and Hinton]{lecun2015deep}
Yann LeCun, Yoshua Bengio, and Geoffrey Hinton.
\newblock Deep learning.
\newblock \emph{Nature}, 521\penalty0 (7553):\penalty0 436--444, may 2015.
\newblock \doi{10.1038/nature14539}.
\newblock URL \url{http://dx.doi.org/10.1038/nature14539}.

\bibitem[Li~Deng(2014)]{deng2014deep}
Dong~Yu Li~Deng.
\newblock \emph{Deep Learning: Methods and Applications}.
\newblock NOW Publishers, May 2014.
\newblock URL
  \url{https://www.microsoft.com/en-us/research/publication/deep-learning-methods-and-applications/}.

\bibitem[Lu et~al.(2014)Lu, Wu, Li, Zhang, Lu, Wang, and
  Zhuang]{lu2014learning}
Xinyan Lu, Fei Wu, Xi~Li, Yin Zhang, Weiming Lu, Donghui Wang, and Yueting
  Zhuang.
\newblock Learning multimodal neural network with ranking examples.
\newblock In \emph{Proceedings of the 22nd ACM international conference on
  Multimedia}, pp.\  985--988. ACM, 2014.

\bibitem[Madjarov et~al.(2012)Madjarov, Kocev, Gjorgjevikj, and
  D{\v{z}}eroski]{madjarov2012extensive}
Gjorgji Madjarov, Dragi Kocev, Dejan Gjorgjevikj, and Sa{\v{s}}o
  D{\v{z}}eroski.
\newblock {An extensive experimental comparison of methods for multi-label
  learning}.
\newblock \emph{Pattern Recognition}, 45\penalty0 (9):\penalty0 3084--3104,
  2012.
\newblock ISSN 0031-3203.
\newblock \doi{http://dx.doi.org/10.1016/j.patcog.2012.03.004}.
\newblock URL
  \url{http://www.sciencedirect.com/science/article/pii/S0031320312001203}.

\bibitem[Makita \& Lenskiy(2016{\natexlab{a}})Makita and
  Lenskiy]{makita2016movie}
Eric Makita and Artem Lenskiy.
\newblock {A movie genre prediction based on Multivariate Bernoulli model and
  genre correlations}.
\newblock \penalty0 (May), mar 2016{\natexlab{a}}.
\newblock URL \url{http://arxiv.org/abs/1604.08608}.

\bibitem[Makita \& Lenskiy(2016{\natexlab{b}})Makita and
  Lenskiy]{makita2016multinomial}
Eric Makita and Artem Lenskiy.
\newblock {A multinomial probabilistic model for movie genre predictions}.
\newblock 2016{\natexlab{b}}.
\newblock URL \url{http://arxiv.org/abs/1603.07849}.

\bibitem[Mao et~al.(2014)Mao, Xu, Yang, Wang, and Yuille]{mao2014explain}
Junhua Mao, Wei Xu, Yi~Yang, Jiang Wang, and Alan~L Yuille.
\newblock Explain images with multimodal recurrent neural networks.
\newblock \emph{arXiv preprint arXiv:1410.1090}, 2014.

\bibitem[Mikolov et~al.(2013{\natexlab{a}})Mikolov, Chen, Corrado, and
  Dean]{mikolov2013efficient}
Tomas Mikolov, Kai Chen, Greg Corrado, and Jeffrey Dean.
\newblock Efficient estimation of word representations in vector space.
\newblock \emph{arXiv preprint arXiv:1301.3781}, 2013{\natexlab{a}}.

\bibitem[Mikolov et~al.(2013{\natexlab{b}})Mikolov, Sutskever, Chen, Corrado,
  and Dean]{mikolov2013distributed}
Tomas Mikolov, Ilya Sutskever, Kai Chen, Greg~S Corrado, and Jeff Dean.
\newblock {Distributed representations of words and phrases and their
  compositionality}.
\newblock In \emph{Advances in Neural Information Processing Systems}, pp.\
  3111--3119, 2013{\natexlab{b}}.

\bibitem[Ngiam et~al.(2011)Ngiam, Khosla, and Kim]{ngiam2011multimodal}
J~Ngiam, A~Khosla, and M~Kim.
\newblock {Multimodal Deep Learning}.
\newblock In \emph{Proceedings of the 28th International Conference on Machine
  Learning (ICML-11)}, pp.\  689----696, 2011.
\newblock URL
  \url{http://ai.stanford.edu/{~}ang/papers/icml11-MultimodalDeepLearning.pdf}.

\bibitem[Norouzi et~al.(2014)Norouzi, Mikolov, Bengio, Singer, Shlens, Frome,
  Corrado, and Dean]{norouzi2014zero}
Mohammad Norouzi, Tomas Mikolov, Samy Bengio, Yoram Singer, Jonathon Shlens,
  Andrea Frome, Greg~S Corrado, and Jeff Dean.
\newblock {Zero-Shot Learning by Convex Combination of Semantic Embeddings}.
\newblock \emph{CoRR}, abs/1312.5, dec 2014.
\newblock URL \url{http://arxiv.org/abs/1312.5650}.

\bibitem[Pais et~al.(2012)Pais, Lambert, Beauchene, Deloule, and
  Ionescu]{pais2012animated}
Gregory Pais, Patrick Lambert, Daniel Beauchene, Francoise Deloule, and Bogdan
  Ionescu.
\newblock {Animated movie genre detection using symbolic fusion of text and
  image descriptors}.
\newblock In \emph{2012 10th International Workshop on Content-Based Multimedia
  Indexing (CBMI)}, number~1, pp.\  1--6. IEEE, jun 2012.
\newblock ISBN 978-1-4673-2369-7.
\newblock \doi{10.1109/CBMI.2012.6269813}.
\newblock URL
  \url{http://ieeexplore.ieee.org/lpdocs/epic03/wrapper.htm?arnumber=6269813}.

\bibitem[Pei et~al.(2013)Pei, Liu, Liu, and Sun]{pei2013unsupervised}
Deli Pei, Huaping Liu, Yulong Liu, and Fuchun Sun.
\newblock {Unsupervised multimodal feature learning for semantic image
  segmentation}.
\newblock In \emph{The 2013 International Joint Conference on Neural Networks
  (IJCNN)}, pp.\  1--6. IEEE, aug 2013.
\newblock ISBN 978-1-4673-6129-3.
\newblock \doi{10.1109/IJCNN.2013.6706748}.
\newblock URL
  \url{http://ieeexplore.ieee.org/lpdocs/epic03/wrapper.htm?arnumber=6706748}.

\bibitem[Shah et~al.(2013)Shah, Motiani, and Patel]{shah2013movie}
Dharak Shah, Saheb Motiani, and Vishrut Patel.
\newblock {Movie Classification Using k-Means and Hierarchical Clustering}.
\newblock Technical report, 2013.

\bibitem[Simonyan \& Zisserman(2014)Simonyan and Zisserman]{simonyan2014very}
Karen Simonyan and Andrew Zisserman.
\newblock Very deep convolutional networks for large-scale image recognition.
\newblock \emph{arXiv preprint arXiv:1409.1556}, 2014.

\bibitem[Socher et~al.(2013)Socher, Ganjoo, Manning, and Ng]{socher2013zero}
Richard Socher, Milind Ganjoo, Christopher~D Manning, and Andrew Ng.
\newblock {Zero-Shot Learning Through Cross-Modal Transfer}.
\newblock In C~J~C Burges, L~Bottou, M~Welling, Z~Ghahramani, and K~Q
  Weinberger (eds.), \emph{Advances in Neural Information Processing Systems
  26}, pp.\  935--943. Curran Associates, Inc., 2013.
\newblock URL
  \url{http://papers.nips.cc/paper/5027-zero-shot-learning-through-cross-modal-transfer.pdf}.

\bibitem[Socher et~al.(2014)Socher, Karpathy, Le, Manning, and
  Ng]{socher2014grounded}
Richard Socher, Andrej Karpathy, Quoc~V Le, Christopher~D Manning, and Andrew~Y
  Ng.
\newblock {Grounded Compositional Semantics for Finding and Describing Images
  with Sentences}.
\newblock \emph{Transactions of the Association for Computational Linguistics
  (TACL)}, 2\penalty0 (April):\penalty0 207--218, 2014.
\newblock URL
  \url{http://nlp.stanford.edu/{~}socherr/SocherLeManningNg{\_}nipsDeepWorkshop2013.pdf}.

\bibitem[Srivastava \& Salakhutdinov(2012)Srivastava and
  Salakhutdinov]{srivastava2012multimodal}
Nitish Srivastava and Ruslan Salakhutdinov.
\newblock {Multimodal Learning with Deep Boltzmann Machines}.
\newblock In F.~Pereira, C.J.C. Burges, L.~Bottou, and K.Q. Weinberger (eds.),
  \emph{Advances in Neural Information Processing Systems 25}, pp.\
  2222--2230. Curran Associates, Inc., 2012.
\newblock URL
  \url{http://papers.nips.cc/paper/4683-multimodal-learning-with-deep-boltzmann-machines.pdf}.

\bibitem[Suk \& Shen(2013)Suk and Shen]{suk2013deep}
Heung~Il Suk and Dinggang Shen.
\newblock {Deep learning-based feature representation for AD/MCI
  classification}.
\newblock In \emph{Lecture Notes in Computer Science (including subseries
  Lecture Notes in Artificial Intelligence and Lecture Notes in
  Bioinformatics)}, volume 8150 LNCS, pp.\  583--590, 2013.
\newblock ISBN 9783642407628.
\newblock \doi{10.1007/978-3-642-40763-5_72}.

\bibitem[Tu et~al.(2014)Tu, Wu, Dai, Jiang, and Xue]{tu2014multimedia}
Jian Tu, Zuxuan Wu, Qi~Dai, Yu-Gang Jiang, and Xiangyang Xue.
\newblock {Challenge Huawei challenge: Fusing multimodal features with deep
  neural networks for Mobile Video Annotation}.
\newblock In \emph{Multimedia and Expo Workshops (ICMEW), 2014 IEEE
  International Conference on}, pp.\  1--6, 2014.
\newblock \doi{10.1109/ICMEW.2014.6890609}.

\bibitem[Van~Merri{\"e}nboer et~al.(2015)Van~Merri{\"e}nboer, Bahdanau,
  Dumoulin, Serdyuk, Warde-Farley, Chorowski, and Bengio]{van2015blocks}
Bart Van~Merri{\"e}nboer, Dzmitry Bahdanau, Vincent Dumoulin, Dmitriy Serdyuk,
  David Warde-Farley, Jan Chorowski, and Yoshua Bengio.
\newblock Blocks and fuel: Frameworks for deep learning.
\newblock \emph{arXiv preprint arXiv:1506.00619}, 2015.

\bibitem[Vinyals et~al.(2015)Vinyals, Toshev, Bengio, and
  Erhan]{vinyals2015show}
Oriol Vinyals, Alexander Toshev, Samy Bengio, and Dumitru Erhan.
\newblock Show and tell: A neural image caption generator.
\newblock In \emph{Proceedings of the IEEE Conference on Computer Vision and
  Pattern Recognition}, pp.\  3156--3164, 2015.

\bibitem[Wu et~al.(2013)Wu, Hoi, Xia, Zhao, Wang, and Miao]{wu2013online}
Pengcheng Wu, Steven~C.H. Hoi, Hao Xia, Peilin Zhao, Dayong Wang, and Chunyan
  Miao.
\newblock {Online multimodal deep similarity learning with application to image
  retrieval}.
\newblock In \emph{Proceedings of the 21st ACM international conference on
  Multimedia - MM '13}, MM '13, pp.\  153--162, New York, New York, USA, 2013.
  ACM Press.
\newblock ISBN 9781450324045.
\newblock \doi{10.1145/2502081.2502112}.
\newblock URL
  \url{http://doi.acm.org/10.1145/2502081.2502112{\%}5Cnhttp://dl.acm.org/citation.cfm?doid=2502081.2502112}.

\bibitem[Xu et~al.(2015)Xu, Ba, Kiros, Cho, Courville, Salakhutdinov, Zemel,
  and Bengio]{xu2015show}
Kelvin Xu, Jimmy Ba, Ryan Kiros, Kyunghyun Cho, Aaron Courville, Ruslan
  Salakhutdinov, Richard~S Zemel, and Yoshua Bengio.
\newblock Show, attend and tell: Neural image caption generation with visual
  attention.
\newblock \emph{arXiv preprint arXiv:1502.03044}, 2\penalty0 (3):\penalty0 5,
  2015.

\bibitem[Yuksel et~al.(2012)Yuksel, Wilson, and Gader]{yuksel2012twenty}
Seniha~Esen Yuksel, Joseph~N Wilson, and Paul~D Gader.
\newblock Twenty years of mixture of experts.
\newblock \emph{IEEE transactions on neural networks and learning systems},
  23\penalty0 (8):\penalty0 1177--1193, 2012.

\bibitem[Zheng et~al.(2014)Zheng, Zhang, and Larochelle]{zheng2014topic}
Yin Zheng, YJ~Zhang, and Hugo Larochelle.
\newblock {Topic Modeling of Multimodal Data: an Autoregressive Approach}.
\newblock In \emph{IEEE Conference on Computer Vision and Pattern Recognition},
  2014.
\newblock ISBN 2011000211.
\newblock URL
  \url{http://www.dmi.usherb.ca/{~}larocheh/publications/ZhengY2014.pdf}.

\end{thebibliography}
\bibliographystyle{iclr2017_conference}

\end{document}